\pdfoutput=1
\documentclass[preprint,12pt,authoryear]{elsarticle}
\usepackage[american]{babel}
\usepackage{natbib} % has a nice set of citation styles and commands
\usepackage{mathtools} % amsmath with fixes and additions
\usepackage{booktabs} % commands to create good-looking tables
\usepackage{tikz} % nice language for creating drawings and diagrams

\usepackage{graphicx}
\usepackage{hyperref}
\usepackage{xcolor}
\usepackage[font=small,labelfont=bf]{caption}
\usepackage{subcaption}
\usepackage{amssymb}
\usepackage{amsmath}
\usepackage{amsfonts}
\usepackage[bottom]{footmisc}
\usepackage{wrapfig}
\usepackage[ruled]{algorithm}
\usepackage{algcompatible}
\usepackage{siunitx}
\usepackage{array}
\usepackage{tabularx}
\usepackage{multirow}
\usepackage{amsmath}
\DeclareMathOperator*{\argmax}{arg\,max}

\usepackage{adjustbox}
\usepackage{amssymb}
\usepackage{bm}
\newcommand{\dataset}{\mathcal{D}}

\newcommand{\pX}{\mathbf{x}}

\definecolor{mypink1}{rgb}{0.858, 0.188, 0.478}

\newcommand{\MRW}[1]{{\color{mypink1}[MW: #1]}}
\newcommand{\ap}[1]{{\color{orange}[AP: #1]}}
\newcommand{\AAA}[1]{{\color{green}[AA: #1]}}
\newcommand{\NP}[1]{{\color{red}[NP: #1]}}

\newcommand{\LL}[1]{{\color{red}[LL: #1]}}
\newcommand{\MK}[1]{{\color{green}[MK: #1]}}
\newcommand{\edit}[1]{{\color{black}#1}}
\newcommand{\sedit}[1]{{\color{black}#1}}

\journal{Artificial Intelligence Journal}

\begin{document}
\newtheorem{myexam}{Example}
\newtheorem{problem}{Problem}
\newtheorem{remark}{Remark}
\newtheorem{proposition}{Proposition}
\newtheorem{proof}{Proof}
\newtheorem{theorem}{Theorem}
\newtheorem{assumption}{Assumption}
\newtheorem{definition}{Definition}
\newcommand{\mw}[1]{\textcolor{red}{[MW: #1]}}

\begin{frontmatter}



\title{Probabilistic Reach-Avoid for\\ Bayesian Neural Networks}


\author{Matthew Wicker$^a$ , Luca Laurenti$^b$, Andrea Patane$^a$,\\ Nicola Paoletti$^c$, Alessandro Abate$^a$, Marta Kwiatkowska$^a$}

\affiliation{organization={Department of Computer Science, University of Oxford, Oxford, UK}}

\affiliation{organization={Delft Center for Systems and Control (DCSC), TU Delft, Delft, Netherlands}}

% \affiliation{organization={Department of Computer Science, Royal Holloway University of London, London, UK}}
\affiliation{organization={Department of Informatics, King's College London, London, UK}}%

\begin{abstract}
%\LL{Let's remember to add changes in a different color}
%\MK{The title does not account for the particular way that BNNs are used in this paper, nor the fact that this is finite horizon}\MRW{I have updated the title to reflect that we study models, but finite-horizon is probably too much detail. }
Model-based reinforcement learning seeks to simultaneously learn the dynamics of an unknown stochastic environment and synthesise an optimal policy for acting in it. Ensuring the safety and robustness of sequential decisions made through a policy in such an environment is a key challenge for policies intended for safety-critical scenarios. In this work, we investigate 
% the problem of computing reach-avoid probabilities for iterative predictions made with Bayesian neural network (BNN) dynamics models. Further, we provide a method for synthesizing an optimal control policy from a given reach-avoid specification and BNN model.
two complementary problems: first, computing reach-avoid probabilities for iterative predictions made with dynamical models, with dynamics described by Bayesian neural network (BNN); second, synthesising control policies that are %\MK{these are not optimal even for the lower bound, because they are obtained by training for which there are no guarantees proven} \MRW{For NNs they are not optimal, but for the numerical solution they are optimal and we can state a discretization error bound, this will be added.}
optimal with respect to a given reach-avoid specification (reaching a ``target'' state, while avoiding a set of ``unsafe'' states) and a learned BNN model. 
% Our solution method leverages bound propagation techniques and backward recursion to compute lower bounds for the probability that sequential actions given by a policy result in reaching a given set of states while avoiding a set of unsafe states according to the BNN model. 
Our solution leverages interval propagation  and backward recursion techniques to compute
%\MK{these bounds can be quite poor, judging from examples} \MRW{Factors contributing to bound tightness are now discussed.}
lower bounds for the probability that a policy's sequence of actions leads to satisfying the reach-avoid specification. 
Such computed lower bounds provide safety certification for the given policy and BNN model. We then introduce %two
control synthesis algorithms to derive policies %\MK{as above, there is no rigorous justification for maximising}\MRW{This has now been addressed}
maximizing said lower bounds on the safety probability. %The two algorithms rely on a numerical solution of the backward recursion and a neural network approximation, respectively. 
We demonstrate the effectiveness of our method on a series of control benchmarks characterized by learned BNN dynamics models. On our most challenging benchmark, compared to purely data-driven policies the optimal synthesis algorithm is able to provide more than a four-fold increase in the number of certifiable states and more than a three-fold increase in the average guaranteed reach-avoid probability. 
\end{abstract}




\end{frontmatter}


\section{Introduction}

{\edit{The capacity of deep learning to approximate complex functions  makes it particularly attractive for inferring process dynamics in control and reinforcement learning problems \citep{schrittwieser2019mastering}.
In safety-critical scenarios where the environment and system state are only partially known or observable (e.g., a robot with noisy actuators/sensors), Bayesian models have recently been investigated as a safer alternative to standard, deterministic, Neural Networks (NNs): the uncertainty estimates of Bayesian models can be propagated through the system decision pipeline to enable safe decision making despite unknown system conditions \citep{mcallister2016data,carbone2020robustness,depeweg2016learning}. 
In particular, \textit{Bayesian Neural Networks} (BNNs) retain the same advantages of NNs (relative to their approximation capabilities) and also enable reasoning about uncertainty in a principled probabilistic manner \citep{neal2012bayesian, murphy2012machine}, making them very well-suited to tackle {safety-critical} problems.}

}

In problems of sequential planning, time-series forecasting, and model-based reinforcement learning, evaluating a model with respect to a control policy (or strategy) 
requires making several predictions that are mutually dependent across time \citep{liang2005bayesian,pilco}.
%In many of these scenarios - e.g, sequential planning, time-series forecasting and control, or model-based RL - in order to evaluate the model (for examples w.r.t.\ a policy or strategy) one often needs to make several predictions correlated across time \citep{liang2005bayesian}.
While multiple models can be learned for each time step, a common setting is for these predictions to be made iteratively by the same machine learning model \citep{huang2020deep}, 
% At each step, the model output is fed back to the model as input for the next step (possibly along with an additional input coming from an external controller) so as to predict, in turn, the output for the following step.  
where the state of the predicted model at each step is a function of the model state at the previous step and possibly of \edit{an action} (from the policy).  
We refer to this setting as \emph{iterative predictions}.

\edit{Unfortunately, performing iterative predictions with BNN models poses several practical issues. In facts,
BNN models output
probability distributions, so that at each successive timestep the BNN needs to be evaluated over a probability distribution, rather than a fixed input point – thus posing the problem of successive
predictions over a stochastic input}.
Even when the \sedit{posterior distribution of the BNN weights is inferred using analytical approximations,
the deep and non-linear structure of the network makes the resulting predictive distribution analytically intractable \citep{neal2012bayesian}. In iterative prediction settings, the problem is compounded and exacerbated by the fact that one would have to evaluate the BNN, sequentially, over a distribution that cannot be computed analytically  \citep{depeweg2016learning}.}
% makes iterative predictions
% with BNNs an analytically intractable problem \citep{neal2012bayesian}.
% To the best of our knowledge, 
Hence, computing sound, formal bounds on the probability of BNN-based iterative predictions remains an open problem. Such bounds would enable one to provide
safety guarantees over a given (or learned) control policy,
which is a necessary precondition before deploying the policy in a real-world environment \citep{polymenakos2020safety,vinogradska2016stability}.

In this paper, we develop a new method for the computation of probabilistic guarantees for iterative predictions with BNNs over \emph{reach-avoid} specifications.  
A reach-avoid specification, also known as constrained reachability \citep{SA13b}, requires that the trajectories of a dynamical system reach a goal/target region over a given (finite) time horizon, whilst avoiding a given set of states that are deemed ``unsafe''. 
Probabilistic reach-avoid is a key property for the formal analysis of stochastic processes \citep{abate2008probabilistic}, underpinning richer temporal logic specifications: %\NP{such as? perhaps this para can be merged with the next} \citep{model_checking,a3c,cauchi2019efficiency}. \MRW{I have merged these two sentances to make things more clear}
its computation is the key component for probabilistic model checking algorithms for various temporal logics such as PCTL, csLTL, or BLTL~\citep{kwiatkowska2007stochastic,cauchi2019efficiency}.  
%\acmt{next sentence needed? which citation?}Note also that probabilistic reach-avoid is a probabilistic generalization of safety guarantees commonly employed to certify model-based reinforcement learning with standard (i.e., non Bayesian) NNs \citep{}.
%
%Because of the probabilistic and non-linear nature of a BNN, exact computation of probabilistic reach-avoid for iterative predictions with BNNs is not feasible. 
%

Even though the exact computation of reach-avoid probabilities for iterative prediction with BNNs is in general not analytically possible, with our method, we can derive a guaranteed (conservative) lower bound 
% for a given policy
% (i.e.\ a pessimistic estimation) 
% can be derived 
by solving a backward iterative problem obtained via a %\MK{also discretisation of action space is hinted at but not discussed in detail} \MRW{This is needed for optimality of the numerical solution, but I agree this is unclear and I have made notes in the synthesis section about what needs to be done}
discretisation of the state space.  
In particular, starting from the final time step and the goal region, we back-propagate the probability lower bounds for each discretised portion of the state space. 
% , starting from the goal region. 
{This backwards reachability approach}  leverages recently developed bound propagation techniques for  BNNs~\citep{wicker2020probabilistic}. In addition to providing guarantees for a given policy, we also 
devise methods to synthesise policies that are maximally certifiable, {i.e., that maximize the lower bound of the reach-avoid probability.} 
% derive schemes for updating policies such that they are maximally certifiable. 
We first describe a numerical solution that, by using dynamic programming, can synthesize policies that are %\MK{don't think they are optimal}
maximally safe. Then, in order to improve the scalability of our approach, we present a method for synthesizing %\MK{no error bound given}
approximately optimal strategies parametrised as a neural network.
\sedit{While our method does not yet scale to state-of-the-art reinforcement learning environments, we are able to verify and synthesise challenging non-linear control case studies.} 

%\MRW{I understand all of your concerns here, the synthesis section must be made more clear wrt the numerical solution proposed.}
%\np{in previous sentence, we need to briefly explain the nature of this discretisation error (and why it's a concern here and not for certification of a fixed policy), and what we mean by first-order}

%\MK{The Intro needs to be revised/restructures and contribution statement improved} 
We validate the effectiveness of our certification and synthesis algorithms on a series of %\MK{are they challenging? single later NNs are used}%\MRW{By control standards these are very challenging scenarios. Consider the henzinger paper and GP control problems cannot even handle the base 6D setting we consider (they can only handle 2 and 3D problems).}
 control benchmarks. %\LL{We need to be careful to not overclaim, as we do not. We really just give guarantees for at most 3 D subsystems for a non-trivial time horizon )}\np{luca, I rephrased to avoid that claim about dimensionality.}
 \edit{Our certification algorithm is able to produce non-trivial safety guarantees for each system that we test. On each proposed benchmark, we also show how our synthesis algorithm results in actions whose safety is significantly more certifiable {than policies derived via deep reinforcement learning}. Specifically, in a challenging planar navigation benchmark, our synthesis method results in policies whose \sedit{certified} safety probabilities %\MK{informal discussion only} %\MRW{Discussion in these sections has now been strengthened}
are eight to nine times higher than those for learned policies. %\sedit{and allow us to provide non-trivial safety guarantees for 20\% of the state space, which could not be certified safe under standard policies.} %\LL{respect to what?}. 
%\textcolor{red}{[in the abstract we also mention `more certifiable states' - recall/mention that point more in detail here.]}
%\LL{Above is a bit unclear. What is a standard policy? } \ap{Honestly I would remove this sentence in blue about certifiable coverage, it is extactly what reviewer 2 is complaining about. I'd focus instead on the average lower bound, which is a hard and tangible guarantee.}

We further investigate how factors like the choice of approximate inference method, BNN architecture, and training methodology affect the quality of the synthesised policy.} 
In summary, this paper makes the following contributions: 
%\LL{5 bullet points are too many...Last two cn be merged}
%
\begin{itemize}
    \item We show how probabilistic reach-avoid for iterative predictions with BNNs can be formulated as the solution of a backward recursion.
    % computation problem, and design an algorithm for the lower bounding of the latter. 
    \item We present an efficient certification framework that produces a lower bound on probabilistic reach-avoid by relying on %\MK{IBP only?}%\MRW{Yes.}
    convex relaxations of the BNN model and said recursive problem definition. %and \LL{Add}
    %\item We discuss how our lower-bound can be used for policy certification. \NP{weak, discussion is not a contribution}
    \item We present schemes for deriving a maximally certified policy (i.e., %\MK{only lower bound?}%\MRW{Corrected.}
    maximizing the lower bound on safety probability) with respect to a BNN and given reach-avoid specification.
    \item We evaluate our methodology on a set of control case studies to provide guarantees for learned and synthesized policies and conduct an empirical investigation %and discussion 
    of model-selection choices and their effect on the %\MK{only informal discussion}%\MRW{Weakened statement.}
    quality of policies synthesised by our method.
\end{itemize}

A previous version of this work \citep{wicker2021certification} has been presented at the thirty-seventh Conference on Uncertainty in Artificial Intelligence. Compared to the conference paper, in this work, we introduce several new contributions. Specifically, compared to \cite{wicker2021certification} we present %\MK{there is only one algorithm, same as in UAI?} 
novel algorithms for the synthesis of control strategies based on both a numerical method and a neural network-based approach.
%provide a new numerical solution method for the synthesis problem, which produces optimally safe policies (as opposed to the approximate synthesis algorithm in \citep{wicker2021certification}). 
Moreover, the experimental evaluation has been consistently extended, to include, among others, an analysis of 
%  we provide the first experiments on the numerical solution to the synthesis problem which is the first time that optimal policies have been synthesized. 
% We also provide greatly extended empirical evaluation and insights into our methods. In particular, this version studies 
the role of approximate inference and NN architecture on safety certification and synthesis, as well as an %\MK{arguable}\MRW{Softened}
in-depth analysis of the scalability of our methods. \edit{Further discussion of related works can be found in Section~\ref{sec:related}.}

\section{Bayesian Neural Networks}

In this work, we consider fully-connected neural network (NN) architectures $f^w:\mathbb{R}^{m}\to\mathbb{R}^n$ parametrised by a vector  $w\in \mathbb{R}^{n_w}$ containing all the weights and biases of the network. Given \edit{a} NN $f^w$ composed by $L$ layers, we denote by $f^{w,1},...,f^{w,L}$ the layers of $f^w$ and we have that $w=\big(\{W_{i}\}_{i=1}^{L}\big) \cup \big(\{b_{i}\}_{i=1}^{L}\big)$, where $W_i$ and $b_i$ represent weights and biases of the $i-$th layer of $f^w.$ %\ap{Matthew can you check? I think the indexes are wrong. In the previous sentence you index the weight matrixes W starting from zero, here you index them starting from one.}\MRW{Checking through now}
For $x\in \mathbb{R}^n$ the output of layer $i\in \{1,...,L\}$ can be explicitly written as $f^{w,i}(x) = a(W_{i}f^{w,{i-1}}(x) + b_{i})$ with $f^{w,1}(x) = a(W_{1}x + b_{1})$, where $a : \mathbb{R} \to \mathbb{R}$ is the activation function. We assume that $a$ is a continuous monotonic function, which holds for the vast majority of activation functions used in practice such as sigmoid, ReLu, and tanh \citep{goodfellow2016deep}. This guarantees that  $f^w$ is a continuous function. %\edit{For further reading on neural networks we reference interested readers to \citep{goodfellow2016deep}. }

Bayesian Neural Networks (BNNs), denoted by $f^{\mathbf{w}}$, extend NNs by placing a prior distribution over the network parameters, $p_{\mathbf{w}}(w)$, with $\mathbf{w}$ being the vector of random variables associated to the parameter vector $w$. 
Given a dataset $ \dataset$, training a  BNN on $\dataset$ requires to compute posterior distribution, $p_{\mathbf{w} } (w \vert \dataset),$ which can be computed via Bayes' rule~\citep{neal2012bayesian}. 
Unfortunately, because of the non-linearity introduced by the neural
network architecture, the computation of the posterior is generally intractable. Hence, various approximation
methods have been studied to perform inference with BNNs
in practice. Among these methods, we consider Hamiltonian
Monte Carlo (HMC) \citep{neal2012bayesian}, and Variational Inference (VI)  \citep{blundell2015weight}. In our experimental evaluation in Section~\ref{sec:res_approx_inf} we employ both HMC and VI.

\paragraph{Hamiltonian Monte Carlo (HMC)} HMC proceeds by defining
a Markov chain whose invariant distribution is $p_{\mathbf{w} } (w \vert \dataset),$ and
relies on Hamiltionian dynamics to speed up the exploration
of the space. Differently from VI discussed below, HMC does not make any \edit{parametric} assumptions on
the form of the posterior distribution and is asymptotically
correct. The result of HMC is a set of samples
that
approximates  $p_{\mathbf{w} } (w \vert \dataset)$. \edit{We refer interested readers to \citep{neal2011mcmc, izmailov2021bayesian} for further details.}

\paragraph{Variational Inference (VI)} VI proceeds by finding a Gaussian approximating distribution $q(w)\sim p_{\mathbf{w} } (w \vert \dataset)$ 
in a trade-off between approximation accuracy and scalability. The core idea is that $q(w)$ depends on some hyperparameters that are then iteratively optimized by minimizing
a divergence measure between $q(w)$ and $p_{\mathbf{w} } (w \vert \dataset)$. Samples
can then be efficiently extracted from $q(w)$. \edit{See \citep{khan2021bayesian, blundell2015weight} for recent developments in variational inference in deep learning.}

\section{Problem Formulation} 
\label{sec:ProbForm}
Given a trained BNN $f^{\mathbf{w}}$ we consider the following discrete-time stochastic process given by iterative predictions of the BNN: 
\begin{align}
\label{Eqn:SystemEqn}
  & \mathbf{x}_k = f^{\mathbf{w}}(\mathbf{x}_{k-1},\mathbf{u}_{k-1})+\mathbf{v}_k, \quad  \mathbf{u}_k = \pi_k(\mathbf{x}_k), \quad k\in \mathbb{N}_{>0},
\end{align}
%\ap{I am not sure this the actual system under consideration. For the way you define the stochastic kernel, I think we }
where  $\mathbf{x}_k$ is a random variable taking values in $\mathbb{R}^n$ modelling the state of System \eqref{Eqn:SystemEqn} at time $k$, $\mathbf{v}_k$ is a random variable modelling an additive noise term with stationary, zero-mean Gaussian distribution $\mathcal{N}(0,\sigma^2\cdot I),$ where $I$ is the identity matrix of size $n \times n$. % where $\sigma^2\cdot I$ is the covariance matrix, with $I$ being the identity matrix of dimension $n$.
 $\mathbf{u}_k$ represents the {action} applied at time $k$, selected from a compact set $\mathcal{U}\subset \mathbb{R}^c$ by 
% from a possibly continuous set of admissible controls $\mathcal{U}$ and is given by 
a (deterministic) feedback Markov strategy (a.k.a. policy, or controller) $\pi:\mathbb{R}^{n}\times \mathbb{N} \to \mathcal{U}$.\footnote{We can limit ourselves to consider deterministic Markov strategies as they are optimal in our setting~\citep{bertsekas2004stochastic,abate2008probabilistic}. Also, in the following, we denote with $\pi$ the time-varying policy described, at each step $k$, by policy $\pi_k:\mathbb{R}^{n} \to \mathcal{U}$.} %We refer to a series of states from this process as a trajectory of the stochastic process or system.  

The model in Eqn.\ \eqref{Eqn:SystemEqn} is commonly employed to represent noisy dynamical models driven by a BNN  and controlled 
by the policy $\pi$ \citep{depeweg2016learning}. %\ap{But we say its a BNN. Why are we now talking of NN? It's confusing imo. Why don't we cite depweg instead that uses BNN?} \LL{We could even just say BNNs and delete last two citations. Up to you guys}
In this setting, $f^{\mathbf{w}}$ defines the transition probabilities of the model and, correspondingly,  
 $p(\bar{x}|(x,u), \mathcal{D})$ is employed to describe the \textit{posterior predictive distribution}, namely the probability density of the model {state} at the next time step being $\bar{x}$, given that the current state and action are $(x,u)$, as:
\begin{align}
    \label{Eqn:PredDIstr}
    p(\bar{x}|(x,u), \mathcal{D})=\int_{\mathbb{R}^{n_w}} \mathcal{N}(\bar{x}\mid f^w(x,u),\sigma^2\cdot I)p_{\mathbf{w}}(w| \mathcal{D}) dw,
\end{align}
where $\mathcal{N}(\cdot \mid f^w(x,u),\sigma^2\cdot I)$ is the Gaussian likelihood induced by noise $\mathbf{v}_k$ %\ap{Maybe you can mention it in the background section that the likelihood function used is the Gaussian.} 
and centered at the NN output \citep{neal2012bayesian}. 

Observe that the posterior predictive distribution induces a probability density function over the state space.
In iterative prediction settings, this implies that at each step the state vector $\mathbf{x}_k$ fed into the BNN is a random variable. Hence, a $N$-step \textit{trajectory} of the dynamic model in Eqn \eqref{Eqn:SystemEqn} is a sequence of states $x_0,...,x_N\in \mathbb{R}^n$ sampled from the predictive distribution. As a consequence, a principled propagation of the BNN uncertainty through consecutive time steps poses the problem of predictions over stochastic inputs. In Section \ref{sec:ReachAvoid} we will tackle this problem for the particular case of reach-avoid properties, by designing a backward computation scheme that starts its calculations from the goal region, and proceeds according to Bellman iterations \citep{bertsekas2004stochastic}.

% \ap{Does this need to be a separate remark? We could just embed this paragraph as it is in the discussion above. We don't have other remarks around, so it really looks ugly, and added like a sticking plaster}

We remark that $p(\bar{x}|(x,u), \mathcal{D})$ is defined by marginalizing over $p_{\mathbf{w}}(w|\mathcal{D})$, 
hence, the particular $p(\bar{x}|(x,u), \mathcal{D})$ depends on the specific approximate inference method employed to estimate the posterior distribution.
As such, the results that we derive are valid w.r.t.\ a specific BNN posterior.

\paragraph{Probability Measure}
For an action $u\in \mathbb{R}^c$, a subset of states $X\subseteq \mathbb{R}^n$ and a starting state $x\in \mathbb{R}^n$, % \MK{is this the standard kernel?}
we call $T(X|x,u)$ the \emph{stochastic kernel} associated (and equivalent \citep{abate2008probabilistic}) to the dynamical model of Equation \eqref{Eqn:SystemEqn}. % induced by the posterior predictive distribution of the BNN, 
Namely, $T(X|x,u)$  describes the one-step transition probability of the model of Eqn.~\eqref{Eqn:SystemEqn} and is defined by integrating the predictive posterior distribution with input $(x,u)$ over $X$, as: 
% \LL{Need to explain the below equation}
\begin{align}
\label{Eqn:KernelFunctionSpace}
    T(X|x,u)=& \int_X p(\bar{x}|(x,u),\mathcal{D}) d \bar{x}. 
\end{align}  
%Note that the above integral is defined over the state space $\mathbb{R}^n$. 
In what follows, it will be convenient at times to work over the space of parameters of the BNN. To do so, we can re-write the stochastic kernel by combining {Equations}~\eqref{Eqn:PredDIstr} and~\eqref{Eqn:KernelFunctionSpace} and applying Fubini's theorem \citep{fubini1907sugli} to switch the integration order, thus obtaining: 
\begin{align} 
  T(X|x,u)= \int_{\mathbb{R}^{n_w}} \left[ \int_X \mathcal{N}(\bar{x}|f^w(x,u),\sigma^2\cdot I) d\bar{x} \right] p_{\mathbf{w}}(w|\mathcal{D}) dw.
   %\label{Eqn:KernelWeightSpace}
\end{align}

From this definition of $T$ it follows that,  under a strategy $\pi$ and for a given initial condition $x_0$, $\mathbf{x}_k$  is a Markov process with a well-defined probability measure $\Pr$ uniquely generated by the stochastic kernel $T$ \citep[Proposition 7.45]{bertsekas2004stochastic} and such that for $X_0,X_k \subseteq \mathbb{R}^n$:
%\footnote{\np{perhaps wrap this paragraph (whole or part of) into a definition}\acmt{needs some re-working: reviewers might be confused looking at this prob measure and stoch kernel, vs. those in the equation above for set X1. }}
%\ap{The notation here needs to be aligned. The state space is a function of the time step here - further down the text we use a subscript notation for that.}\acmt{these are just sets Xk within the state space}
\begin{align*}
& \Pr[\pX_0\in X_0] =  \mathbf{1}_{X_0}(x_0), \\
& \Pr[\pX_k\in X_k  |\pX_{k-1}=x, \pi] =  T^{}(X_k|x,
% \pi(x,k-1)
\pi_{k-1}(x)),
\end{align*}
%\subsection{Problem Formulation}
%\acmt{we should align throughout the paper the use of initial conditions (set $X_0$): whilst is it OK to use if in the next equations, I do not think it is strictly necessary for the statement of Problem 2 and in general for the Reach-Avoid problem (see also comment by NP in Def. 1)}\LL{How would you modify it? Would you just ask the bound to hold for all $x\in \mathrm{S}\cup\mathrm{G}?$}
%\acmt{I think the above formula with the Xk sets is fine, but I'd eliminate the initial set Xo from the ReachAvoid problems.}
where $\mathbf{1}_{X_0}$ is the indicator function (that is, $1$ if $x \subseteq X_0$ and $0$ otherwise). Having a definition of $\Pr$ allows one to make probabilistic statements over the stochastic model in Eqn \eqref{Eqn:SystemEqn}.

\sedit{
\begin{remark}
Note that, as is common in the literature \citep{depeweg2016learning}, according to the definition of the probability measure $\Pr$ we marginalise over the posterior distribution at each time step. Consequently, according to our modelling framework, the weights of the BNN are not kept fixed during each trajectory, but we re-sample from $\mathbf{w}$ at each time step. 
\end{remark}
%{Throughout the paper, we will assume that each step of the system transitions according to the posterior predictive distribution as has been investigated in prior works \cite{pilco, pilconn}. Other predictive strategies such as maxing subsequent predictions based on a fixed random sample from the posterior would lead to a different dynamical system than what is considered in this work.}
%

%Note that, as common in Bayesian settings  \citep{depeweg2016learning},  the weights of the BNN are not kept fixed during a trajectory of System \eqref{Eqn:SystemEqn}, but we marginalise over the posterior distribution at each step. This is motivated by the fact that we use a BNN to learn a one-step transition model, where the predictive posterior distribution of the BNN is used to model the transition of the state from step $k$ to step $k+1$. 

}

\subsection{Problem Statements}

We consider two problems concerning, respectively, the certification and the control of dynamical systems modelled by BNNs. We first consider safety certification with respect to probabilistic reach-avoid specifications. That is, we seek to compute the probability that from a given state, under a selected control policy, an agent navigates to the goal region without encountering any unsafe states. 
% a problem that we solve by computing sound lower bounds on the probability that from a given state, under a selected control policy, an agent navigates to the goal region without encountering any unsafe states. 
Next, we consider the formal synthesis of policies that maximise this probability and thus attain maximal certifiable safety.

\begin{problem}[Computation of Probabilistic Reach-Avoid]
\label{Prob:verification}
Given a strategy $\pi$, a goal region $\mathrm{G}\subseteq \mathbb{R}^n$, a finite-time horizon $[0,N]\subseteq \mathbb{N},$  and a safe set $\mathrm{S}\subseteq \mathbb{R}^n$ such that $\mathrm{G}\cap \mathrm{S}= \emptyset $, compute for any $x_0 \in \mathrm{G}\cup \mathrm{S}$
%\LL{update notation without $\rho$}
% \footnote{\np{also, here do we want to give in input $x_0$? the thing is that the init state of the trajectory is already defined, so, making $P_{reach}$ parametric in $x_0$ makes little sense. Or we can make the trajectory parametric on $x_0$. Further, I guess we mean here $x_0$ (a concrete value)}\LL{Let's assume a fixed initial condition. (otherwise I will have to update the proof )}}
\begin{align}
 \nonumber    & P_{reach}(\mathrm{G},\mathrm{S},x_0,[0,N]|\pi) = \\
 & \hspace{1cm} \Pr\big[\exists k \in [0,N], \pX_k \in \mathrm{G}\, \wedge \forall 0 \leq k' < k, \pX_{k'} \in \mathrm{S} \mid \pX_0=x_0,\pi\big].\label{Eqn:ProbForm}
\end{align}
%\acmt{slightly confused about $x(k)$ vs $x_k$ notation}
\end{problem}

\paragraph{Outline of the Approach}  
In Section \ref{sec:ReachAvoid} we show how $P_{reach}(\mathrm{G},\mathrm{S},x_0,[0,N]|\pi)$ %\ap{For clarity, either put [0,N] or describe what k is in one sentence, also I wouldn't use k since it is itelf used in Equation 5 that defines Preach}\MRW{I agree that it should just be [0,N] here for clarity}
can be formulated as the solution of a backward iterative computational procedure, where the uncertainty of the BNN is propagated backward in time, starting from the goal region. Our approach allows us to compute a sound lower bound on  $P_{reach}$, thus guaranteeing that $\pX_k,$ as defined in Eqn \eqref{Eqn:SystemEqn}, satisfies the specification with a given probability. This is achieved by extending existing lower bounding techniques developed to certify  BNNs  \citep{wicker2020probabilistic} and applying these at each propagation step through the BNN.

% Each propagation through the BNN can be done by extending existing lower bounding techniques developed to certify  BNNs  \citep{wicker2020probabilistic}, which allow us to compute a sound lower bound on  $P_{reach}$, thus guaranteeing that the process $\pX_k$ satisfies the specification with a given probability.  %We will show that such a formulation of $P_{reach}$ has \AAA{two} main advantages.
%Firstly, it allows us to define techniques for certification of BNNs to compute a sound lower bound on  $P_{reach}$, thus guaranteeing that the process $\pX_k$ satisfies the specification with a given probability. \AAA{secondly?}

% \subsection{Strategy Synthesis}

Note that, in Problem \ref{Prob:verification}, the strategy $\pi$ is provided, and the goal is to quantify the probability with which the %\MK{trajectories not defined} \MRW{Corrected}
trajectories of $\mathbf{x}_k$ satisfy the given specification.  In Problem~\ref{Prob:Syntesis} below, we expand the previous problem and seek to synthesise a controller $\pi$ that maximizes $P_{reach}$. The general formulation of this optimization is given below.

\begin{problem}[Strategy Synthesis for Probabilistic Reach-Avoid] 
\label{Prob:Syntesis}
For an initial state $x_0 \in \mathrm{G} \cup \mathrm{S}$, and a finite time horizon $N$, find a strategy $\pi^{*}:\mathbb{R}^{ n}\times \mathbb{R}_{ \geq 0}\to\mathbb{R}^c$ such that
\begin{align}
  \pi^{*} = \argmax_{\pi} P_{reach}(\mathrm{G},\mathrm{S},x_0,[0,N]\mid \pi).
\end{align}
\end{problem}
%Obtaining an optimal policy, $\pi^{*}$, in the unconstrained space of policies, $\mathbb{R}^{ n}\times \mathbb{R}_{ \geq 0}\to\mathbb{R}^c$, is generally intractable. Therefore, unlike in Problem 1, we place some moderate restrictions on the classes of functions that our policy can come from.
In Section~\ref{Sec:Syntesis}, we will provide specific schemes for synthesizing optimal strategies when $\pi$ is either a look-up table or a deterministic neural network. 

\paragraph{Outline of the Approach} To solve this problem, we notice that the backward iterative procedure outlined to solve Problem 1 has a substructure such that dynamic programming will allow us to compute optimal actions for each state that we verify, thus producing an optimal policy with respect to the given posterior and reach-avoid specification. {With low-dimensional or discrete action spaces, we can then derive a tabular policy by 
%discretising the action space and 
solving the resulting dynamic programming problem.}
%\MK{approx error not analysed} \MRW{Added}
%\LL{I am not sure the following sentence is correct}\np{I think there is an approximation error induced by the action discretisation, but this is not $1/o$ (it's hard to quantify how much we underestimate the true optimal probability due to this error, so I would skip the $1/o$ sentence)}\MRW{Yes, you are both correct, that optimally here is w.r.t. the maximally certifiably safe policy, not the maximally safe policy. Maybe it is best to leave it out. Another option is to add the phrase I have at the end of the sentence to clear it up a bit. Happy either way.}
%The error in our solution to this optimization problem is dependent on the discretization of the action space. For example partitioning a single dimensional action (bounded on the unit interval) into $o$ uniform, equidistant partitions we can ensure that the found optimal action is at most $1/o$ \edit{away from the maximally certifiably safe action}. More complex numerical schemes can be implemented to achieve different error rates \citep{higham2002accuracy}.
For higher-dimensional action spaces instead, in Section \ref{sec:NeuralNetworksSynthesis} we consider (generalising) policies represented as neural networks. %\LL{Next sentence is unclear. A reviewer had a doubt on this, so we need to clarify and elaborate as we do in the rebuttal} 
\section{Methodology}\label{sec:method}

In this section, we illustrate the methodology used to compute lower bounds on the reach-avoid probability, as described in Problem 1. 
% that a trajectory from a given BNN and policy respects a reach-avoid specification. 
% \AAA{[statement not very clear:] Here we deal on a lower level of abstraction than the definitions themselves, opting to make clear any necessary theoretical assumptions} while leaving specific computational details to the algorithms and experimental sections to follow. \MRW{I agree, thanks}
We begin by encoding the reach-avoid probability through a sequence of value functions. %\edit{i.e., the Bellman equations}. %We then introduce lower bounds for such value functions, defined w.r.t.\ some discretization of the state space. Such lower bounds will be used to certify the probabilistic behavior of the system.  
% casting safety as a computation of value functions. 
% We then provide the certification of probabilistic behavior by lower bounding the value functions encoding a reach-avoid property.  %Finally, we discuss how this certification can also results in \AAA{bounds on confidence intervals} that can be used for real-time decision support. \MRW{We removed this uncertainty problem statement so this needs to be removed} 

\subsection{Certifying Reach-Avoid Specifications}\label{sec:ReachAvoid}

{We begin by showing that} $P_{reach}(\mathrm{G},\mathrm{S},x,[k,N]|\pi)$ can be obtained as the solution of a backward iterative procedure, which allows to compute a lower bound on its value. %The first step will be to \MK{what is meant by this?} define value functions appropriate for a model with BNN dynamics. 
In particular, given a time $0\leq k<N$ and a strategy $\pi,$ consider the value functions $V_k^{\pi}:\mathbb{R}^n\to [0,1]$, recursively defined as  
%\ap{I remember in meetings you used to call this the value function? Is that standard nomenclature? If it is it is worth mentioning it in here maybe for clarity?}
\begin{align}
    &\hspace*{-0.1cm}\nonumber V_N^{\pi}(x)=\mathbf{1}_\mathrm{G}(x), \\
    &\hspace*{-0.1cm}V_k^{\pi}(x)=\mathbf{1}_\mathrm{G}(x)+\mathbf{1}_\mathrm{S}(x){\hspace*{-0.1cm} \int \hspace*{-0.1cm} V^{\pi}_{k+1}(\bar{x}) p\big(\bar{x} | (x, \pi_{k}(x)),\mathcal{D}\big) d\bar{x}}. %=\mathbf{1}_\mathrm{G}(x)+\mathbf{1}_\mathrm{S}(x)\mathbb{E}_{T^{} (\bar{x} \mid x)}[ V^{\pi}_{k+1}(\bar{x})]
    \label{Eqn:ExactValueFunc}
\end{align}
%\ap{The descritise formula is given in the weight space - it might be good to give also this in weight space for consistency?}
%\ap{I think below is a bit confusing. Maybe we can write something of the form: ``Intuitively, $V_k^{\pi}$ reformulates the calculation of Preach through backward computations. Starting from the goal region $G$, which has ofc probability $1$ of reaching the goal assigned; the computation proceed backwards for each state $x$ by combining the transition probabilities coming from the predictive distribution, and the reach probabilities already computed up until that point. This is similar to what is done in dynamical rpogramming [some references] and is formally stated in the following Proposition that is proved in the supplementary.'' or something like this? Also we call it ``backward'' - but is that standard nomenclature? To me it doesn't look backward, it looks like a whole mess in which the computation in every $x$ depend on every other state $\bar{x}$}
Intuitively, $V_k^{\pi}$ is computed starting from the goal region $\mathrm{G}$ at $k=N$, where it is initialised at value $1$.
The computation proceeds backwards at each state $x$, by combining the current values with the transition probabilities from Eqn~\eqref{Eqn:SystemEqn}.  
%it is not equal to zero on those states either in $\mathrm{G}$ or such that can reach $\mathrm{G}$, according to Equation \eqref{Eqn:SystemEqn},
%with non-zero probability within the next $N-k$ steps, while dwelling within the safe set $\mathrm{S}$.
The following proposition, proved {inductively over time} in the Supplementary Material, guarantees that $ V_0^{\pi}(x)$ is indeed equal to $P_{reach}(\mathrm{G},\mathrm{S},x,[0,N]|\pi)$. %\mw{Why is this only specifically for k=0? } \AAA{If I understand correctly, the val fcn at 0 encompasses the probability of the reach-avoid spec over the entire time horizon. Looks OK.} 
\begin{proposition}
\label{th:CorrectnessBack}
For $0\leq k\leq N$ and $x_0 \in \mathrm{G}\cup\mathrm{S},$ it holds that 
$$  
P_{reach}(\mathrm{G},\mathrm{S},x_0,[k,N]|\pi)= V_k^{\pi}(x).
$$
\end{proposition}
%\np{do we need a sketch of proof for the above?}\LL{Added that the proof is in the Supplementary}
\noindent
The backward recursion in Eqn~\eqref{Eqn:ExactValueFunc} does not generally admit a solution in closed-form, as it would require integrating over the BNN posterior predictive distribution, which is in general analytically intractable. %\edit{Despite the intractability, we can formally reason about the posterior predictive distribution of a BNN over discrete portions of the input space by using recently proposed convex relaxations \citep{wicker2020probabilistic}.}
In the following section, we present a computational scheme utilizing convex relaxations to lower bound $P_{reach}$. % that can be used to formally reason about a given BNN and certify a given strategy.   

\begin{figure}[h]
    \centering
    \includegraphics[width=0.75\textwidth]{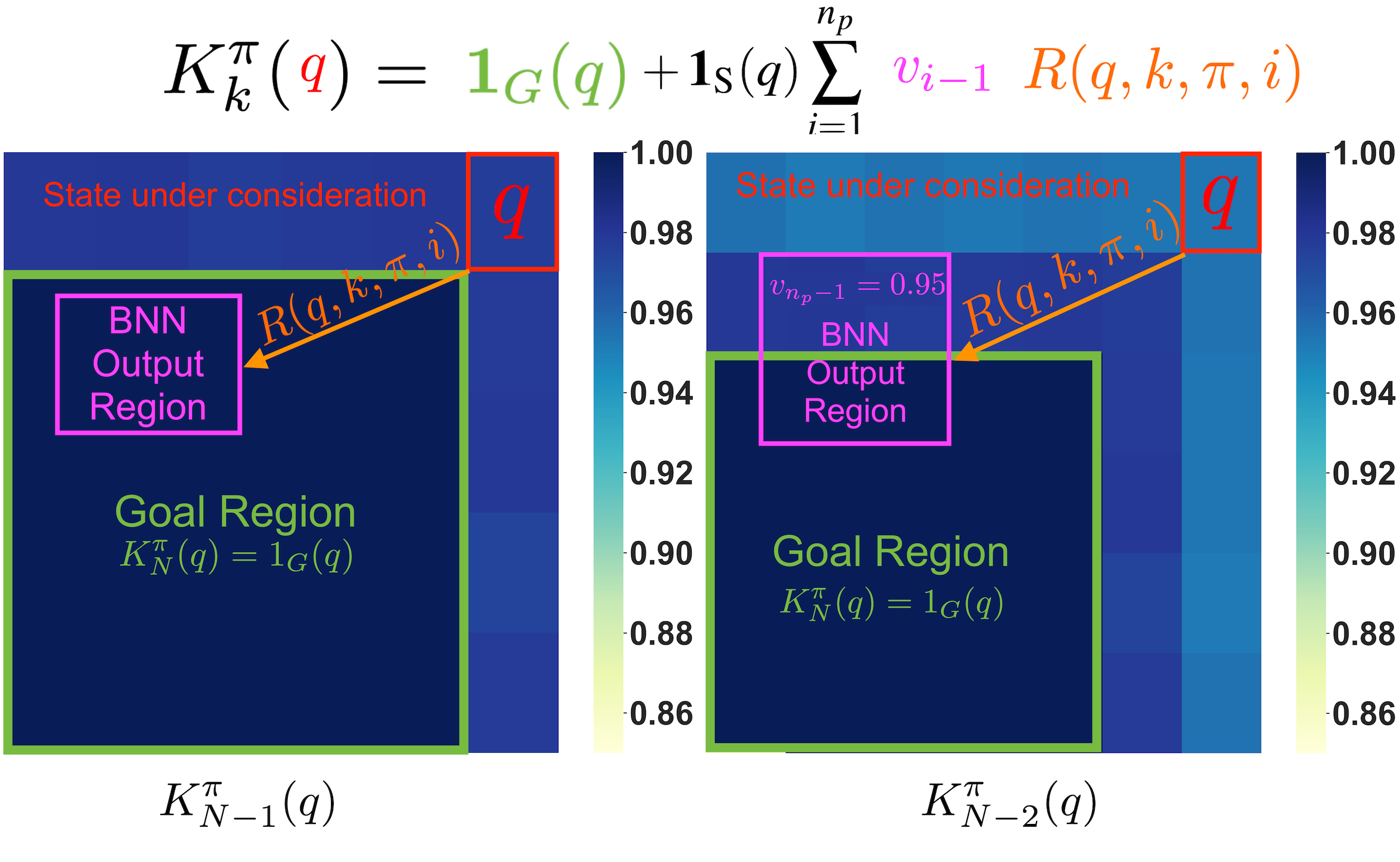}
    \caption{
    Examples of functions $K^{\pi}_{N-1}$ (left) and $K^{\pi}_{N-2}$ (right), which are lower bounds of $V^{\pi}_k$. 
    % for any $0\leq k\leq N$. % The heatmaps illustrate the values of $K^{\pi}_{k}$ for a two-dimensional state space. 
    % We illustrate our approach for the computation of functions $K^{\pi}_{k}$ that are a guaranteed lower bound of $V^{\pi}_k$ for any $0\leq l\leq N$. 
    %\MK{Informal, needs polishing}
     On the left, we consider the first step of our backward algorithm, where we compute $K^{\pi}_{N-2}(q)$ by computing the probability that $\mathbf{x}_{N}\in \mathrm{G}$ given that $\mathbf{x}_{N-1}\in q.$  On the right, we consider the subsequent step. We outline the state we want to verify in red and the goal region in green. With the orange arrow, we represent the 0.95 transition probability of the BNN dynamical model, and in pink we represent the worst-case probabilities spanned by the BNN output. On top, we show where each of these key terms comes into play in Eqn~\eqref{Eqn:ValueFuncDiscre}.  }\label{fig:notation_visualization}
\end{figure}

%
%In the following section we show that this formulation of $P_{reach}$ enables us to formulate an  algorithm for computing a lower bound for this quantity. %In particular, we show how a  lower bound can be efficiently computed, and (see Section \ref{Sec:Syntesis}) how this also allows to synthesise robust strategies for $V_{0}^{\pi}$. \LL{This is a repetition}
%
\subsection{Lower Bound on $P_{reach}$}\label{sec:LowerBound}

We develop a computational approach based on the %\MK{fix terminology: sometimes use abstract state, or partition state, or partitioned state}\MRW{Done}
discretisation of the state space, \edit{which allows convex relaxation methods such as \citep{wicker2020probabilistic} to be used}. %\MK{and actions?}\MRW{This is only for synthesis. Actions are given by the policy in this case. This will be more clear once the synthesis section is updated}
The proposed computational approach is illustrated in Figure \ref{fig:notation_visualization} and formalized in Section \ref{sec:Lowbound}. 
Let $Q=\{q_1,...,q_{n_q} \}$ be a partition of $\mathrm{S} \cup \mathrm{G}$ in $n_q$ regions and denote with  $z:\mathbb{R}^n\to Q$ the function that associates to a state in $\mathbb{R}^n$ the corresponding partitioned state in $Q$. 
For each $0\leq k \leq N$ we iteratively build a  set of functions $K^{\pi}_{k}: Q \rightarrow [0,1]$ such that for all $x\in \mathrm{G}\cup \mathrm{S}$ we have that  $K^{\pi}_{k}(z(x))\leq V^{\pi}_{k}(x)$.
Intuitively, $K^{\pi}_{k}$ provides a lower bound for the value functions on the computation of $P_{reach}$.
%super-level sets of the form 
%$$ \mathcal{L}_i^k=\{ {x}\in \mathbb{R}^n \, : \,  V^{\pi}_{k}(x) \geq v_{i+1}  \}. $$
%Then, we will have that $ P_{reach}(\mathrm{G},\mathrm{S},x,[0,N]\mid \pi)\geq \bar{v}$, where $\bar{v}= \max_{i \text{ s.t. }x\in \mathcal{L}_i^0} v_i$ \ap{I don't understand this last statement}.

%\MK{Too informal, if this is a method then write as algorithm}\MRW{This is written algorithmically. See Algorithm 1}
The functions $K^{\pi}_{k}$ are obtained by propagating backward the BNN predictions from time $N$, where we set $K^{\pi}_{N}(q)=\mathbf{1}_{\mathrm{G}}(q)$, with $\mathbf{1}_{\mathrm{G}}(q)$ being the indicator function (that is, $1$ if $q\subseteq \mathrm{G}$ and $0$ otherwise).  Then, for each $k<N$, we first discretize the set of possible probabilities in $n_p$ sub-intervals $0=v_0\leq v_1\leq...\leq v_{n_p}=1$.
Hence, for any $q\in Q$ and probability interval $[v_i,v_{i+1}]$, one can compute a lower bound, $R(q,k,\pi,i)$, on the probability that, starting from any state in $q$ at time $k$, we reach in the next step a region that has probability $\in [v_i,v_{i+1}]$ of %\MK{where is convergence proof?}\MRW{This is not the sense in which the word convergence was used here. Changed for clarity.}
safely reaching the goal region.
%\np{next sentence is unclear. shall we say, ``for any $q\in Q$ and any probability interval $[v_i,v_{i+1}]$, we compute a lower bound $R(q,k,\pi,i)$ on the probability that starting from a state in $q$ at time $k$,  we reach in the next step a region that has probability $\in [v_i,v_{i+1}]$ of converging to the goal region.'' and then we can directly introduce the equations of (8), remarking that that the sum of the products between the $v_i$s and such $R$s is a lower bound to the integral of (7)?} 
%we propagate $q$ through the predictive posterior distribution obtaining $n_p$ distinct regions where  $\mathbf{x}_{k+1}$ will be with high probability given that $\mathbf{x}_{k}\in q$ (in Figure \ref{fig:notation_visualization} we show only the last BNN Output Region, which only contains states $q'$ such that $K^{\pi}_{k+1}(q')\in [v_{n_p-1},1]$). 
The resulting values are used to build $K^{\pi}_{k}$ (as we will detail in Eqn \eqref{Eqn:ValueFuncDiscre}). For a given $q\subset \mathrm{S}$, $K^{\pi}_{k}(q)$ is obtained as the sum over $i$ of $R(q,k,\pi,i)$ multiplied by $v_{i-1}$, i.e.,  the lower value that $K^{\pi}_{k+1}$ obtains in all the states of the $i-th$ region. Note that the discretisation of the probability values does not have to be uniform, but can be adaptive for each $q\in Q$. A heuristic for picking the value of thresholds $v_i$ will be given in Algorithm \ref{alg:backwardsreachability}. %For the ease of a simpler notation we will omit the dependence of each $v_i$ from $q$.
In what follows, we formalise the intuition behind this computational procedure.
%We should note that in the Eqn reported on the top of Figure \ref{fig:notation_visualization} the sum is over $n_p$. In fact, the assumption is that there are $n_p$ BNN output regions such that only the i-th region states $q'$ such that $K^{\pi}_{k+1}(q')\in [v_i,v_{i+1}]$.
%\LL{Need to be careful to not count the same probability twice and to polish a bit. Hopefully, this will help tho }

\subsection{Lower Bounding of the Value Functions} \label{sec:Lowbound}
%\MK{Need to define $\gamma$, etc, and give intuition, especially for the weight sets; the role of $\eta$ is unclear and its impact on precision} \MRW{Added a sentance on $\eta$, but $\gamma$ is just a placeholder variable derived from $\epsilon$ (and thus $\eta$)}
For a given strategy $\pi$, \edit{we} consider a constant $\eta\in (0,1)$ and $\epsilon=\sqrt{2\sigma^2}\text{erf}^{-1}(\eta)$, which are used to bound the value of the noise, $\mathbf{v}_k$, at any given time. Intuitively, $\eta$ represents the proportion of observational error we consider.\footnote{The threshold is such that it holds that $Pr(|\mathbf{v}_k^{(i)}|\leq \epsilon)=\eta$. In the experiments of Section \ref{sec:experiments} we select $\eta=0.99$.} 
Then, for $0\leq k<N$, $K^{\pi}_k:Q \to [0,1]$ are defined recursively as follows:  
\begin{align}
    &K^{\pi}_N(q)=\mathbf{1}_{\mathrm{G}}(q), \label{Eqn:ValueFuncDiscreTrivial}\\
    &K^{\pi}_k(q)=\mathbf{1}_{\mathrm{G}}(q) + \mathbf{1}_{\mathrm{S}}(q) \sum_{i=1}^{n_p}v_{i-1}R(q,k,\pi(q),i)  ,% 
    \label{Eqn:ValueFuncDiscre}
\end{align}
where
\begin{align}
&R(q,k,\pi(q),i)= \eta^n \int_{H^{q,\pi,\epsilon}_{k,i}}p_{\mathbf{w}}(w|\mathcal{D})   dw, \label{eq:def_of_R}\\ 
&  H^{q,\pi,\epsilon}_{k,i}=\{ w \in \mathbb{R}^{n_w} | \, \forall x\in q, \forall \gamma \in [-\epsilon,\epsilon]^n, \text{ it holds that: } \nonumber \\
  &\hspace{0.3cm}  v_{i-1} \leq  K_{k+1}^{\pi}(q') \leq v_{i}, \text{ with $q'=z(f^w(x,\pi_{k}(x))+\gamma)$} \}. \nonumber
  \end{align}
%and  $\mathbf{1}_{\mathrm{G}}(q)=\inf_{x\in q}\mathbf{1}_{\mathrm{G}}(x)$.%  $\mathbf{1}_{\mathrm{S}}(q)=\inf_{x\in q}\mathbf{1}_{\mathrm{S}}(x)$, $p_{\mathbf{w}}(w|\mathcal{D})$ is the posterior distribution of the parameters of the BNN, and 
%\LL{We also have the dependence on $\epsilon$ in $H$. Need to find a better notation if we can.}
The key component for the above backward recursion is %\MK{R used differently} \MRW{I agree $\pi$ shouldnt take an argument here, corrected.}
$R(q,k,\pi,i) $, which bounds the probability that, starting from $q$ at time $k$, we have that $\mathbf{x}_{k+1}$ will be in a region $q'$ such that $K_{k+1}^{\pi}(q')\in [v_i,v_{i+1}]$.
{By definition, the set} $H^{q,\pi,\epsilon}_{k,i}$ defines the weights for which {the BNN maps all states covered by $q$ into the goal states given action $\pi(q)$}. {Given this, it is clear that} integration of the posterior $p_{\mathbf{w}}(w|\mathcal{D})$ over the  $H^{q,\pi,\epsilon}_{k,i}$ will return the probability mass of system \eqref{Eqn:SystemEqn} transitioning from $q$ to $q'$ with probability in $[v_i,v_{i+1}]$ in one time step.
The computation of Eqn \eqref{Eqn:ValueFuncDiscre} then reduces to computing the set of weights $H^{q,\pi,\epsilon}_{k,i}$, which we call the \textit{projecting weight set}.
 % Note also that sets of weights $H^{q,\pi,\epsilon}_{k,i}$ and $H^{q,\pi,\epsilon}_{k,j}$ for $i\neq j$ are not overlapping, except for at most a set of most measure zero. 
A method to compute a safe under-approximation $ \bar{H}\subseteq  H^{q,\pi,\epsilon}_{k,i}$ is discussed below. %\ap{I would say something ``Notice that in order to compute this you have to compute H - a method for that is given in the following section'' }.
Before describing that, we analyze the correctness of the above recursion. %\acmt{Reviewers will soon wonder about the computability of the sets above: I would add a short note that anticipates the discussion in the next subsection.}
%The following theorem guarantees that $K^{\pi}_k$ is a sound lower bound of $V^{\pi}_k.$
\begin{theorem}
\label{th:VerificationLoerBound}
Given $x\in \mathbb{R}^n$, for any  $k\in \{0,...,N\}$ and $q = z(x)$,  assume that $H^{q,\pi,\epsilon}_{k,i} \cap H^{q,\pi,\epsilon}_{k,j}=\emptyset$ for $i\neq j.$ Then:
$$ \inf_{x\in q} V_{k}^{\pi}(x)\geq  K^{\pi}_k(q).$$
\end{theorem}
A proof of Theorem \ref{th:VerificationLoerBound} is given in the Supplementary Material.
Note that the assumption on the null intersection between different projecting weight sets required in Theorem \ref{th:VerificationLoerBound} can always be enforced by taking their intersection and complement.

\subsection{Computation of Projecting Weight Set  %\AAA{title perhaps not very useful - improve?}
}
\label{sec:WeightSetsComputation}

%\ap{I think this section atm, is mostly tailored toward VI - but we mostly do HMC in the experiments. Bits and pieces need to be adapted and made consistent.}\MRW{Updated. I think a minor adaption is all that was needed. }

Theorem \ref{th:VerificationLoerBound} allows us to compute a safe lower bound to Problem \ref{Prob:verification}, by relying on an abstraction of the state space, that is, through the computation of $K^{\pi}_{0}(q)$. This can be evaluated once the projecting set of weight values $H^{q,\pi,\epsilon}_{k,i}$ associated to $[v_{i-1},v_i]$ is known.\footnote{In the case of Gaussian VI the integral of Equation \eqref{eq:def_of_R} can be computed in terms of the $\mathit{erf}$ function, whereas more generally Monte Carlo or numerical integration techniques can be used.} Unfortunately, direct computation of $H^{q,\pi,\epsilon}_{k,i}$ is intractable.
Nevertheless, a method for its lower bounding was developed by \cite{wicker2020probabilistic} in the context of adversarial perturbations for 
%\MK{there is also a one-step method in Lechner et al} \MRW{They use a different technique which we done. We should discuss that we only use wicker2020 and say similar schemese like Lechner and deepmind could also be used}
one-step BNN predictions, and can be directly adapted to our settings.

The idea is that an under approximation $\bar{H} \subseteq H^{q,\pi,\epsilon}_{k,i}$ is built by sampling weight boxes of the shape $\hat{H} = [w^L,w^U]$, according to the posterior, and checking whether: 
\begin{align}
    v_{i-1}\leq K_{k+1}^\pi (z(f^w({x},\pi_{k}(x)  ) +  \gamma ) )\leq v_{i},  \nonumber \\  \forall x \in q, \, \forall w \in \hat{H}, \, \forall \gamma \in [-\epsilon,\epsilon]^n. \label{eq:one_step_condition}
\end{align}
Finally, $\bar{H}$ is built as a %\MK{not disjoint in algo}\MRW{You are right. This is corrected.}
disjoint union of boxes $\hat{H}$  satisfying the above condition. \edit{For a full discussion of the details of this method we refer interested readers to \citep{wicker2020probabilistic}.}
In order to apply this method to our setting, we propagate the abstract state $q$ through the policy function $\pi_{k}(x)$, so as to obtain a bounding box $\widehat{\Pi} = [\pi^L,\pi^U]$ such that $\pi^L \leq \pi_{k}(x) \leq \pi^U$ for all $x \in q$. 
In the experiments, this bounding is only necessary when $\pi_{k}(x)$ is given by \edit{an} NN controller, for which bound propagation of NNs can be used for the computation of $\widehat{\Pi}$ \citep{gowal2018effectiveness,gehr2018ai2}.

%\MK{The paper needs to be self-contained, so need to state formally what you rely on}\MRW{Done, thanks!}
The results of Proposition 2 and Proposition 3 from \cite{wicker2020probabilistic} can then be used to propagate $q$, $\widehat{\Pi}$ and $\hat{H}$ through the BNN. For discrete posteriors (e.g., those resulting from HMC) one can use {the method described by} \cite{gowal2018effectiveness} (Equations 6 and 7). Propagation of  $q$, $\widehat{\Pi}$ amounts to using these method to compute values $f^{L}_{q,\epsilon,k}$ and $f^{U}_{q,\epsilon,k}$ such that, for all  $x \in q, \gamma \in [-\epsilon,\epsilon]^n, w \in \hat{H}$, it holds that:
\begin{align}\label{eq:ibp_final_results}
  f^{L}_{q,\epsilon,k} \leq f^w({x},\pi_{k}(x)  ) +  \gamma  \leq f^{U}_{q,\epsilon,k}.
\end{align}
Furthermore, $f^{L}_{q,\epsilon,k}$ and $f^{U}_{q,\epsilon,k}$ are differentiable w.r.t.\ to the input vector \citep{gowal2018effectiveness, wicker2021bayesian}.

Finally, the two bounding values can be used to check whether or not the condition in Eqn \eqref{eq:one_step_condition} is satisfied, by simply checking whether $ [   f^{L}_{q,\epsilon,k},  f^{U}_{q,\epsilon,k} ] $  propagated through $K_{k+1}^{\pi}$ is within $[v_i,v_{i+1}]$. We highlight that computing this probability is equivalent to a conservative estimate of $R(q, k, \pi, i)$. %Now that we have the necessary ingredients, in the following we describe our algorithm for the lower bounding of $P_{reach}$.

\subsection{Probabilistic Reach-Avoid Algorithm }

In Algorithm~\ref{alg:backwardsreachability} we summarize our approach for computing a lower bound for Problem \ref{Prob:verification}. 
For simplicity of presentation, we consider the case $n_p=2$, (i.e., we partition the range of probabilities in just two intervals $[0,v_1],$ $[v_1,1]$ - the case $n_p>2$ follows similarly). 
The algorithm proceeds by first initializing the reach-avoid probability for the partitioned states $q$ inside the goal region $G$ to $1$, as per Eqn \eqref{Eqn:ValueFuncDiscreTrivial}.
Then, for each of the $N$ time steps and for each one of the remaining abstract states $q$, in line $4$ we set the threshold probability $v_1$ equal to the maximum value that $K^{\pi}$ attains at the next time step over the states in the neighbourhood of $q$ (which we capture with a hyper-parameter $\rho_x > 0$). 
We found this %\MK{is it discussed?} 
heuristic for the choice of $v_1$ to work well in practice (notice that the obtained bound is formal irrespective of the choice of $v_1$, and different choices could potentially be explored). 
We then proceed in the computation of Eqn \eqref{Eqn:ValueFuncDiscre}. 
This computation is performed in lines $5$--$14$.
First, we initialise to the null set the current under-approximation of the projecting weight set, $\bar{H}$.
We then sample $n_s$ weights boxes $\hat{H}$ by sampling weights from the posterior, and expanding them with a margin $\rho_w$ heuristically selected (lines 6-8).
Then, for each of these sets, we first propagate the state $q$, policy function, and weight set $\bar{H}$ to build a box $\bar{X}$ according to Eqn \eqref{eq:ibp_final_results} (line 9), which is then accepted or rejected based on the value that $K^{\pi}$ at the next time step attains in states in $\bar{X}$ (lines 10-12). $K^{\pi}_{N-i}(q)$ is then computed in line 14 by integrating $p_{\mathbf{w}}(w|\mathcal{D})$ over the union of the accepted sets of weights.

\begin{algorithm} 
%\algsetup{linenosize=\small}
\caption{Probabilistic Reach-Avoid for BNNs}\label{alg:backwardsreachability}\small
\textbf{Input:} BNN model $f^{\mathbf{w}}$, safe region $\mathrm{S}$, goal region $\mathrm{G}$, discretization  $Q$ of $\mathrm{S} \cup \mathrm{G}$, time horizon $N$, neural controller $\pi$, number of BNN samples $n_s$, weight margin $\rho_w$, state space margin $\rho_x$ \\
\textbf{Output:} Lower bound on $V^{\pi}$ \\
\vspace*{-0.35cm}
%\ap{The notation of this algorithm is to be adjusted to that of the paper}
\begin{algorithmic}[1]
\STATE For all $0\leq k \leq N$ set $K^\pi_k(q)=1$ iff $q\subseteq \mathrm{G}$ and $0$ otherwise
\FOR {$k\gets N$ to $1$} 
    \FOR{$q \in Q \setminus \mathrm{G}$}
        \STATE $v_1 \gets \max_{x\in [q-\rho_x,q+\rho_x]} K^{\pi}_{k+1}(z(x)) $
        %\STATE $[\overline{y},$ \underbar{$y$}] $\gets [-\infty, \infty]$ \COMMENT{over-approximation of output set}
        \STATE $\bar{H} \gets \emptyset$ $\quad$ \text{\# $\bar{H}$ is the set of safe weights}
        \FOR{desired number of samples, $n_s$}
            \STATE $w' \sim P(w | \mathcal{D})$
            \STATE $\hat{H} \gets [w'-\rho_w, w'+\rho_w]$
            %\STATE \# Propagation according to $(\text{Eqn \eqref{eq:ibp_final_results})}$
            \STATE $\bar{X} \gets [ f^{L}_{q,\epsilon,k}, f^{U}_{q,\epsilon,k}] \quad$ \text{\# Computed according to Eqn~\eqref{eq:ibp_final_results}}%\gets \text{Prop.}( q,\,\pi,\,\hat{H},\, \gamma)$
            \IF{$\min_{x\in \bar{X}}K^{\pi}_{k+1}(z(x))\geq v_1$}
                \STATE $ \bar{H} \gets \bar{H} \bigcup \hat{H}$
            \ENDIF
        \ENDFOR
        \STATE \text{Ensure $H_i \cap H_j = \emptyset \quad \forall H_i, H_j \in \bar{H}$}
        \STATE $K^\pi_{k}(q) = v_1\cdot \eta^n\int_{\bar{H}}p_{\mathbf{w}}(w|\mathcal{D})dw$ \; (Eqn \eqref{Eqn:ValueFuncDiscre})
    \ENDFOR
\ENDFOR
\STATE \textbf{return} $K^\pi$
\end{algorithmic}
\end{algorithm}

\section{Strategy Synthesis}\label{Sec:Syntesis}
%In this section, we then show how the backwards computation for bounding value functions admits a substructure that allows us to numerically synthesise provably safe policies, solving Problem 2. In the case of parameterized policy representations, e.g., NNs, we also provide a method for synthesizing safer actions.
We now focus on synthesising a strategy that maximizes our lower bound on $P_{reach}$, thus solving Problem 2. Notice that, while no global optimality claim can be made about the strategy that we obtain, maximising the lower bound guarantees that the true reach-avoid probability will still be greater than the improved bound obtained after the maximisation.
\begin{definition}\label{def:max-cert}
A strategy $\pi^*$ is called maximally certified (max-cert), w.r.t. the discretised value function $K^\pi$,  if and only if, for all $x \in  \mathrm{G}\cup\mathrm{S}$, it satisfies  
$$  
K_{0}^{\pi^*}(z(x))=\sup_{\pi} K_{0}^{\pi}(z(x)),
$$
that is, the strategy $\pi^*$ maximises the lower bound of $P_{reach}$.
\end{definition}
%\LL{We may also want to show explicitly that such a policy extsts, even tho may not be needed for the conference version. It should suffice the set of admissible controls is a compact set }
\noindent
It follows that, if $K_{0}^{\pi^*}(z(x))>1-\delta$ for all $x \in \mathrm{G} \cup \mathrm{S}$, then the max-cert strategy $\pi^*$ is a solution of Problem 2.
Note that a max-cert strategy is guaranteed to exist when the set of admissible controls $\mathcal{U}$ is compact \cite[Lemma 3.1]{bertsekas2004stochastic}, as we assume in this work.  
In the next theorem, we show that a max-cert strategy can be computed via dynamic programming with a backward recursion similar to that of Eqn \eqref{Eqn:ValueFuncDiscre}.
\begin{theorem}
\label{Th:Syntesis}
For  $0\leq k<N$ and $0=v_0<...<v_{n_p}=1,$ define the functions $K^*_{k}:\mathbb{R}^n\to [0,1]$  recursively as follows
\begin{align*}
    &K^*_{k}(q)=\sup_{u\in \mathcal{U}} \big( \mathbf{1}_\mathrm{G}(q) + \mathbf{1}_\mathrm{S}(q) \sum_{i=1}^{n_p}v_{i}R(q,k,u,i)\big), 
\end{align*}
where $R(q,k,u,i)$ and $H^{q,u,\epsilon}_{k,i}$ are defined as in Eqn \eqref{eq:def_of_R}.
%\begin{align*}
%&R(q,k,u,i)= \eta^n \int_{H^{q,\pi,\epsilon}_{k,i}}p_{\mathbf{w}}(w|\mathcal{D})   dw,\\ 
%&  H^{q,\pi,\epsilon}_{k,i}=\{ w \in \mathbb{R}^{n_w} : \, \forall x\in q, \forall \gamma \in [-\epsilon,\epsilon]^n, \text{ it holds that }\\
%  &\hspace{0.5cm}  v_{i-1} \leq  K_{k+1}^{u}(q') \leq v_{i}, \text{ where $q'=z(f^w(x,u)+\gamma)$} \},
%  \end{align*}
If $\pi^*$ is s.t.\ $K^*_0=K^{\pi^*}_0$, then $\pi^*$ is a max-cert strategy. 
Furthermore, for any $x$, it holds that $K^{\pi^*}_0(z(x))\leq P_{reach}( \mathrm{G},\mathrm{S},[0,N],x | \pi^*)$. 
\end{theorem}
Theorem \ref{Th:Syntesis} is a direct consequence of the Bellman principle of optimality \cite[Theorem 2]{abate2008probabilistic} and it guarantees that for each state $q\in \mathrm{S}$  and time $k$, we have that $\pi^*(q,k)=\text{arg}\max_{u\in \mathcal{U}}  \sum_{i=1}^{n_p}v_{i}R(q,k,u,i).$

%Theorem \ref{Th:Syntesis} allows one to recursively compute a max-cert strategy, by selecting at each time step the action that maximizes $K_k$.
%

In Algorithm \ref{alg:actionoptimization} we present a numerical scheme based on Theorem \ref{Th:Syntesis} to find a max-cert policy $\pi^*$.  Note that the optimization problem required to be solved at each time step state, i.e. $\text{arg}\max_{u\in \mathcal{U}}  \sum_{i=1}^{n_p}v_{i}R(q,k,u,i),$  is non-convex. Hence, in Algorithm \ref{alg:actionoptimization}, in Line 1, we start by partitioning the action space $\mathcal{U}$. Then, in Lines 4--15 for each action in the partition we estimate the expectation of $K^*_{k+1}$ starting from $q$ via $n_s$ samples taken from the BNN posterior (250 in all our experiments). Finally, in Lines 11--14 we keep track of the action maximising $K^*_{k+1}$.

%discretizing the 

%$R(q,k,u,i)$ depends on $k$ we have that the resulting $\pi^*$ will generally depend on time, i.e., $\pi^*(\cdot,k)\neq \pi^*(\cdot,k+1)$.

%for each state $q\in \mathrm{S}$  at time $k$, we have that $\pi^*(q,k)=\text{arg}\max_{u\in \mathcal{U}}  \sum_{i=1}^{n_p}v_{i}R(q,k,u,i).$ Note that as  $R(q,k,u,i)$ depends on $k$ we have that the resulting $\pi^*$ will generally depend on time, i.e., $\pi^*(\cdot,k)\neq \pi^*(\cdot,k+1)$.

The described approach for synthesis, while optimal in the limit of an infinitesimal discretization of $\mathcal{U}$, may become infeasible for large state and action spaces. % In fact, it requires to solve a non-convex optimization problem for every $q \in Q$ and $u \in \mathcal{U}$. 
%As $\sum_{i=1}^{n_p}v_{i}R(q,k,u,i)$ is in general non-convex with respect to $u$, its solution will in general require numerical schemes that discretize $\mathcal{U}$.
%\MK{where discussed?}
% as a possibly non-convex optimization problem needs to be solved for every state-action pair, i.e., for every $q \in Q$ and every $u \in \mathcal{U}$.  
As a consequence, in the next subsection, we also consider when  $\pi$ is parametrised by a neural network and thus can serve as a function over a larger (even infinite) state space. 
Specifically, we show how a set of neural controllers, one for each time step, can be trained in order to maximize probabilistic reach-avoid via Theorem~\ref{Th:Syntesis}.  %to modify a loss \MK{can you really claim optimal?}\LL{If we use a NN as a controller, of course we cannot claim optimality, but I do not think we indeed claim optimality for that case} that can lead to actions with improved safety. 
In Section~\ref{sec:experiments} we empirically investigate both controller strategies. %primarily study the numerical solution to this problem and investigate NN synthesized strategies for high-dimensional control problems.

%\subsection{Numerical Algorithm for Strategy Synthesis}

%In Algorithm~\ref{alg:actionoptimization}, we present an outline of the numerical approach to computing optimal actions. We start by taking a placeholder optimal action to be the one given by the supplied policy, $\pi$. We then consider each action from a uniformly discredited set of actions (line 5), and for each we estimate a lower bound on its safety (lines 7-12). The estimate of safety is computed by propagating the abstract state $q$ along with the action $u_i$ through the BNN, and then checking the worst-case lower bound (from $K^{\pi}$ in respective the output region. This estimate is a statistical estimate of the probabilistic reach-avoid lower bound whose error can be bounded via concentration inequalities. In practice, we use 250 BNN samples to estimate this quantity. Once the worst-case quantity has been estimated, we check if this action maximizes safety (line 13) and if we store it for comparison with the remaining actions. In order to synthesize an optimal policy, one performs this procedure on line 4 of Algorithm~\ref{alg:backwardsreachability} and updates the passed value for $K^{\pi}$ as it is computed. Then, for each $q$, one saves the optimal action, $u^{*}$, in a look-up table which represents a numerically optimal policy. 

\begin{algorithm} 
%\algsetup{linenosize=\small}
\caption{Numerical Synthesis of Action for region $q$ at time $k$}\label{alg:actionoptimization}\small
\textbf{Input:} BNN model $f^{\mathbf{w}}$, safe region $\mathrm{S}$, goal region $\mathrm{G}$, action space $\mathcal{U}$, abstract state $q\in Q$, controller $\pi$, number of BNN samples $n_s$\\
\textbf{Output:} Action maximizing $K^{\pi}$ \\

\vspace*{-0.35cm}
%\ap{The notation of this algorithm is to be adjusted to that of the paper}
\begin{algorithmic}[1]
\STATE $\Upsilon \gets $ middle points of each region in a partition of $\mathcal{U}$
%\STATE $u^{*} \gets \pi(q)$
%\STATE \text{\# Estimated safety of action}
\STATE $\kappa^{*} \gets 0$
\STATE $u^{*} \gets 0$
\FOR{$u \in \Upsilon$}
    \STATE $\hat{\kappa} \gets 0$
    \FOR{$j$ from $0$ to $n_s$}
            \STATE $w' \sim P(w | \mathcal{D})$
            %\STATE \# Propagation according to $(\text{Eqn \eqref{eq:ibp_final_results})}$ and $u_i$ and $w'$
            \STATE $\bar{X} \gets [ f^{L}_{q,\epsilon,k}, f^{U}_{q,\epsilon,k}] $ \# Computed for $q$ and $u$ via  $(\text{Eqn \eqref{eq:ibp_final_results})}$
            \STATE $\hat{\kappa} = \hat{\kappa} + \frac{\min_{x\in \bar{X} }K^{*}_{k+1}(z(x))}{n_s}$
    \ENDFOR
    \IF{$\hat{\kappa} > \kappa^{*}$}
        \STATE $\kappa^{*} \gets \hat{\kappa}$
        \STATE $u^{*} \gets u$
    \ENDIF
\ENDFOR
\STATE \textbf{return} $u^{*}$
\end{algorithmic}
\end{algorithm}

\subsection{An Approach for Strategy Synthesis Based on Neural Networks}
\label{sec:NeuralNetworksSynthesis}

%\MK{This is too informal, does not clarify the representation of initial policy, needs an algorithm with properly identified inputs/outputs/assumptions and correctness statement}\LL{Modified}
In this subsection we show how we can train a set of  NN policies $\pi_{0},...,\pi_{N-1}: \mathbb{R}^n \to \mathcal{U}$  such that at each time step $k$, $\pi_{k}$ approximately solves the dynamic programming equation in Theorem \ref{Th:Syntesis}. Note that, because of the approximate nature of the NN training, the resulting neural policies will necessarily be sub-optimal, but have the potential to scale to larger and more complex systems, compared to the approach presented in Algorithm \ref{alg:actionoptimization}.

At time $k$ we start with an initial set of parameters (weights and biases) $\theta_k$ for policy $\pi_{k}$. These parameters can either be initialized to ${\theta_{k+1}}$, the parameters synthesised at the previous time step of the value iteration for $\pi_{k+1}$, or to a policy employed to collect the data to train the BNN as in \cite{pilconn}, or simply selected at random. \edit{ In our implementation where no previous policy is available, we start with a randomly initialized NN, and then at time $k$ we set our initial neural policy with that obtained at time $k+1$.} %\LL{Modified, check.} \ap{Can we use something that is not H for the initial time, since H is the interval set everywhere else?}\MRW{Resolved I think.}
We then employ a scheme to learn a ``safer'' set of parameters via backpropagation. In particular, we first define the following loss function penalizing policy parameters that lead to an unsafe behavior for an ensemble of NNs sampled from the BNN  posterior distribution: 
\begin{align}\label{eq:loss}
\mathcal{L}(x,\theta_k)=-\alpha || \sum_{w \in \bar{W}} f^w(x,\pi_{k}(x)) -\mathbf{A}_k ||_2 +  (1-\alpha) || \sum_{w \in \bar{W}} f^w(x,\pi_{k}(x)) -\mathbf{R}_k ||_2, 
\end{align}
where $\bar{W}$ are a set of parameters independently sampled from the BNN posterior $p_{\mathbf{w} } (w \vert \dataset),$ for a probability threshold $0\leq p_t \leq 1,$ $\mathbf{A}_k=\{x : K^{\pi_{{k+1}}}_{k+1}(x)\geq  p_t \}$ and $\mathbf{R}_k=\{x : K^{\pi_{{k+1}}}_{k+1}(x)\leq 1- p_t \}$ are the sets of states for which the probability of satisfying the specification at time $k+1$ is respectively greater than $p_t$ and smaller than $1-p_t$. For $X\subset \mathbb{R}^n,$ $|| x -X ||_2=\inf_{\bar{x}\in X} || x - \bar{x}||_2 $ is the standard $L_2$ distance of a point from a set, and $0\leq \alpha \leq 1$ is a parameter taken to be 0.25 in our experiments, that weights between reaching the goal and staying away from ``bad" states. Intuitively, the first term in $\mathcal{L}(x,\theta_k)$ enforces $\theta_k$ that leads to high values of $K^{\pi_{{k+1}}}_{k+1}$, while the second term penalizes parameter sets that lead to small values of this quantity. %\LL{Would not be enough to just consider the first part of the loss?} \MRW{Perhaps, I think the idea between the second term is to break ties. Consider you have two directions that are, in the next time-step, equally safe. The one closer to the unsafe states is less preferable. Perhaps in practice we could have removed the second term, though.}

$\mathcal{L}(x,\theta_k)$ only considers the behaviour of the dynamical system of Equation \eqref{Eqn:SystemEqn} starting from initial state $x$. Then, in order to also enforce robustness in a neighbourhood of initial states around $x$, similarly to  the adversarial training case \citep{madry2017towards}, we consider the robust loss 
\begin{align}
\label{eqn:RobustLoss}
\bar{\mathcal{L}}(x,\theta_k)=\max_{x':||x-x' ||_2 \leq \epsilon} \mathcal{L}(x,\theta_k),
\end{align}
%where $\epsilon$ is a parameter taken to be 0.025. \LL{This we should say in the experiments not in here.}
Note that by employing Eqn \eqref{eq:ibp_final_results} we obtain a differentiable upper bound of $\bar{\mathcal{L}}(x,\theta_k)$, which can be employed for training $\theta_k$. %The resulting training procedure is reported in Algorithm \ref{}. 

\subsection{\edit{Discussion on the Algorithms}}
%\LL{IF this section is all new, do highlight it all. ALso, I am a bit confused about the purpose of this Section. We present two different algorithms for control. Which one are we talking about? Also, while the discussion about complexity makes sense in here, the discussion about how overapproximate it is may make more sense in the experiments or in the Conclusions}

\edit{In this section we provide further discussion of our proposed algorithms including the complexity and the various sources of approximation that may lead to looser guarantees. To frame this discussion, we start by highlighting the complexity and approximation introduced by the chosen bound propagation technique shared by both of the algorithms. We then proceed to discuss how discretisation choices made with respect to the state-space, the weight-space, and the observational noise, practically affect the tightness of our probability bounds for both algorithms, and finally how the action-space discretisation affects our synthesis algorithm.  %\np{``affect the guarantees offered by our approach'' should this be ``affect the tightness of our probability bounds'' (we don't want to give the impression that guarantees are compromised)}\MRW{That is a very good point, thank you!}

\paragraph{Bound Propagation Techniques} %A key component of the bounds presented in this work are derived from bound propagation techniques for Bayesian neural networks \citep{wicker2020probabilistic, wicker2021certification, berrada2021make} which are used to compute the value of the $R$ function. %Crucially, the $R$ function returns a sound lower bound on the one-step probability of safety. 
Given that there are currently no methods for BNN certification that are both sound and complete \citep{wicker2020probabilistic, wicker2021certification, berrada2021make}, the evaluation of the $R$ function will always introduce some approximation error. While it is difficult to characterize this error in general, it is known that for deeper networks, BNN certification methods introduce more approximation than shallow networks \citep{wicker2021adversarial}. The recently developed bounds from \cite{berrada2021make}, have been shown to be tighter than the IBP and LBP approaches from \cite{wicker2020probabilistic} at the cost of computation complexity that is exponential in the number of dimensions of the state-space. In contrast, each iteration of the interval bound propagation method proposed in and \cite{wicker2020probabilistic} requires the computational complexity of four forward passes through the neural network architecture. %\ap{Why four? That is not clear to me. I guess you mean that it is a parameter how many you sample, and in practice already with 4 it works well enough.} \MRW{I see your confusion. Check the updated sentence.}
%The complexity and approximations below will be discussed in terms of how many times one must evaluate the $R$ function.

\paragraph{Discretization Error and Complexity}

While our formulation supports any form of state space discretisation, we can assume for simplicity that each dimension of the $n$-dimensional state-space is broken into $m$ equal-sized abstract states. This implies that certification of the system requires us to evaluate the $R$ function $\mathcal{O}\big(N(m^{n})\big)$ many times, where $N$ is the time horizon we would like to verify. Given that $n$ is fixed, the user has control over $m$, the size of each abstract state. For large abstract states, small $m$, one introduces more approximation as the $R$ function must account for all possible behaviors in the abstract state. For small abstract states, large $m$, there is much less approximation, but considerably larger runtime.  % In practice, the systems we study take roughly 6 hours to certify with a parallelized implementation over 90 CPU cores.
%Our discretization of the action-space only affects our synthesis algorithm in low-dimensional  settings (i.e., where $n$ is small). \np{not clear why this is the case, what happens if $n$ is large instead?} \MRW{Youre right this is a non-sensical sentence. Not sure what I meant here.}
Assume the $c$-dimensional  action space is broken into $t$ equal portions at each dimension, then the computational complexity of the algorithm becomes $\mathcal{O}\big(t^{c}N(m^{n})\big)$ as each of the $m^{n}$ states must be evaluated $t^{c}$-many times to determine the approximately optimal action. As with the state-space discretization, larger $t$ will lead to a more-optimal action choice but requires greater computational time.  %In practice, we find that using a statistical estimate of the certification probability leads to an equally safe synthesized policy and admits a complexity of $\mathcal{O}\big(t^{c} + K(m^{n})\big)$. \np{is the statistical estimate explained in previous sections? if not, explain here or omit. also, the reason for this complexity is not fully clear, at least to me.} In practice, the systems we study take roughly 8 hours to synthesize (and certify) with a parallelized implementation over 90 CPU cores.

}
\section{Experiments}\label{sec:experiments}

% \ap{Following reviewer 2 comments I would change the explanation of Cert. Coverage from "we are able to certify" to "we are able to compute non-zero certificates". Or something on these lines. I put it as a general comments because it is mentioned several times.}\MRW{Maybe someone else has edited this, but it seems that how you want it described is how we describe it. Let me know if I am being silly and overlooking something.}

We provide an empirical analysis of our BNN certification and policy synthesis methods.
We begin by providing details on the experimental setting in Section~\ref{sec:exp_settings}.
% , including specifics of the agent dynamics for the problems we study, the three different obstacle environments we will analyse, and the learning algorithm by which we set the initial policy. 
We then analyse the performance of our certification approach on synthesized policies in Section~\ref{sec:synsthesis_exp}. Next, in Section~\ref{sec:res_approx_inf}, we discuss how the choice of the BNN inference algorithm affects synthesis and certification results. 
% Next, we visualize our results on strategy synthesis for each obstacle layout and discuss the benefits of employing a maximally certifiable policy. 
Finally, in Section \ref{sec:scalability} we study how our method scales with larger neural network architectures and in higher-dimensional control settings. %, in particular in a system with  and find that our methods for certification can work on systems with up to 48 dimensions and that we can synthesize safer NN policies. 

\subsection{Experimental Setting}\label{sec:exp_settings}
%\ap{Learned policy, initial policy, learnt policy and maximally ceritfied policy, optimal policy, synthesized policy. Pick one name  each and use it consistently.}\MRW{I had a quick scan and think this is good now, thanks to you if you already corrected it!}

We consider a planar control task consisting of a point-mass agent navigating through various obstacle layouts.
The point-mass agent is described by four dimensions, two encoding position information and two encoding velocity \citep{Astrom08}. To control the agent there are two continuous action dimensions, which represent the force applied on the point-mass in each of the two planar directions.
The task of the agent is to navigate to a goal region while avoiding various obstacle layouts. 
The knowledge supplied to the agent about the environment is the locations of the goal and obstacles. 
% \edit{Given an action and a state, the underlying dynamics are detailed fully in Appendix~\ref{appendix:agentdynamics}. }
\edit{The full set of equations describing the agent dynamics is given in~\ref{appendix:agentdynamics}.}
\edit{In our experiments,} we analyse three obstacle layouts of varying difficulty, which we name \texttt{v1}, \texttt{v2} and \texttt{Zigzag} - visualized in the left column of Figure~\ref{fig:SynthesisFigure}. 
%\ap{don't understand why we refer to figure 3 here and not figure 2 which is closer.} % \ap{Can we make a copy of the left column of figure 3 (without the trajectories) to help visualise them here, rather than having to scroll three pages down?}. % - so to evaluate the effect on the bounds and policies obtained. 
Obstacle layout \texttt{v1} places an obstacle directly between the agent's initial position and the goal, forcing the agent to navigate its way around it. Obstacle layout \texttt{v2} extends this setting by adding two further obstacles that block off one side of the state space. Finally, scenario \texttt{Zigzag}  has 5 interleaving triangles and requires the agent to navigate around them in order to reach the goal. %Each of these are visualized in Figure~\ref{fig:LearnedSystems}.

In order to learn an initial policy to solve the task, we employ the episodic learning framework \edit{described in \cite{gal2016improving}.} %\LL{Add citation} \ap{Yes, we should make clear that we are not making things up, but that we are existing a method from the literature. This was one of the confusion of reviewer 1}
\edit{This consists of iteratively collecting data from deploying our learned policy, updating the BNN dynamics model to the new observations, and updating our policy.} When collecting data, we start by randomly sampling state-action pairs and observing their resulting state according to the ground-truth dynamics. After this initial sample, all future observations from the ground-truth environment are obtained from deploying our learned policy. The initial policy is set by assigning a random action to each abstract state. This is equivalent to tabular policy representations in standard reinforcement learning \citep{sutton}. We additionally discuss neural network policies in Section~\ref{sec:scalability}.  %representation we rely on discretisation of the action space, 
\edit{Actions in the policy are updated by performing gradient descent on a sum of discounted rewards over a pre-specified finite horizon. The reward of an action is taken to be the $\ell_2$ distance moved towards the goal region penalized by the $\ell_2$ proximity to obstacles as is done in \citep{sutton, gal2016improving}.}
%\np{I would replace the text ``This consists of\ldots obtained from deploying our learned policy.'' with ``At each episode, we collect data using the current policy, and use the data to update the BNN dynamics model and the policy''.}\MRW{I have made the change, thanks!}
%
For the learning of the BNN,
%Given a dataset of state-action pairs and their resulting state, 
we perform approximate Bayesian inference over the neural network parameters. For our primary investigation, we select \edit{an} NN architecture with a single fully connected hidden layer comprising 50 hidden units, %\ap{It would be good if we had a citation here to justify this NN size}%\MRW{There is no need to justify this choice. It is studied in later sections. No other method does this so it is not taking inspiration from anywhere.}
and learn the parameters via Hamiltonian Monte Carlo (HMC). Larger neural network architectures are considered in Section~\ref{sec:scalability}, while results for variational approximate inference are given in Section~\ref{sec:res_approx_inf}.  %We primarily study BNNs inferred with Hamiltonian Monte Carlo (HMC) but also study variational posterior approximations and the effect of this choice. Regardless of approximate inference choice, given that the BNN faithfully replicates the observed dynamics, we can employ it to update our policy. 

Unless otherwise specified, in performing certification and synthesis we employ abstract states \sedit{spanning a width of $0.02$ around each position dimension and $0.08$ around each velocity dimension. Velocity values are clipped to the range $[-0.5, 0.1]$.} When performing optimal synthesis, we discretise the two action dimensions for the point-mass problem into 100 possible vectors which uniformly cover the continuous space of actions $[-1, 1]$. When running our backward reachability scheme, at each state, we test all 100 action vectors and take the action that maximizes our lower bound to be the policy action at that state, thus giving us the locally optimal action within the given discretisation. Further experimental details are presented in \ref{sec:appendixdetails} and code to reproduce all results in this paper can be found at \href{https://github.com/matthewwicker/BNNReachAvoid}{https://github.com/matthewwicker/BNNReachAvoid}.

\edit{The computational times for each system were roughly equivalent. This is to be expected given that each has the same state space. The following average times are reported for a parallel implementation of our algorithm run on 90 logical CPU cores across 4 Intel Core
Xeon 6230 clocked at 2.10GHz. Training of the initial policy and BNN model %\np{do we mean training of the initial policy and BNN model?}\MRW{Good point}
takes in the order of 10 minutes, 6 hours for the certification with a horizon of 50 time steps, and 8 hours for synthesis.}

\begin{figure}[H]
\centering
\includegraphics[width=1.0\textwidth]{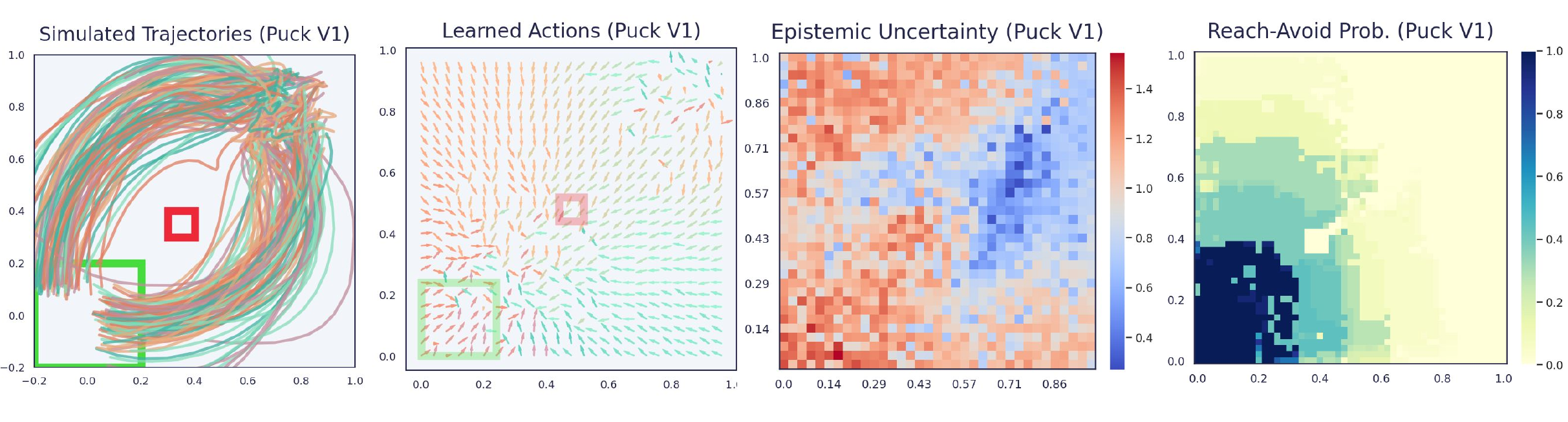}
\includegraphics[width=1.0\textwidth]{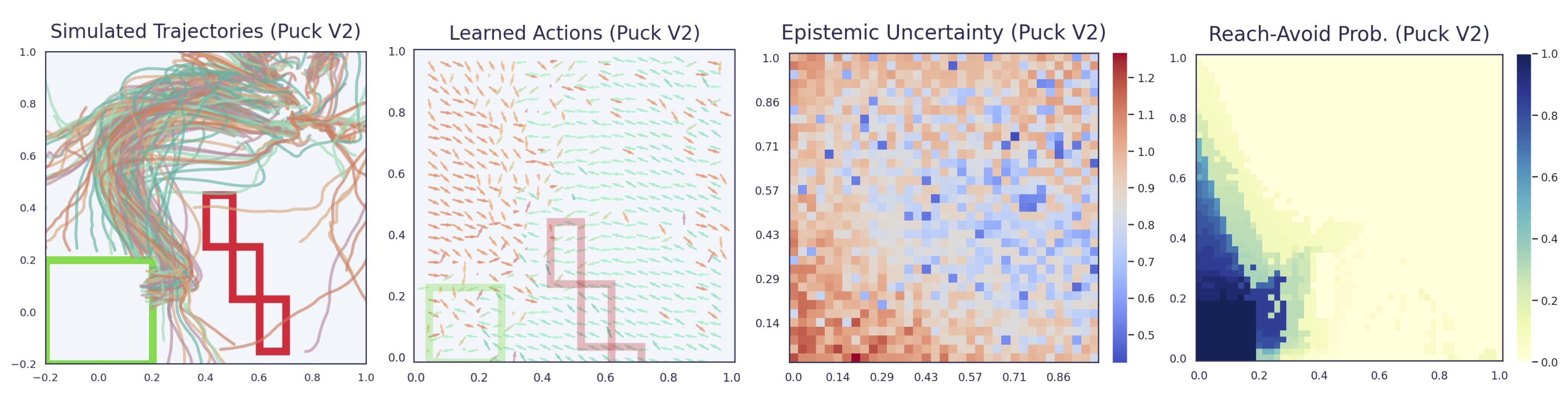}
\includegraphics[width=1.0\textwidth]{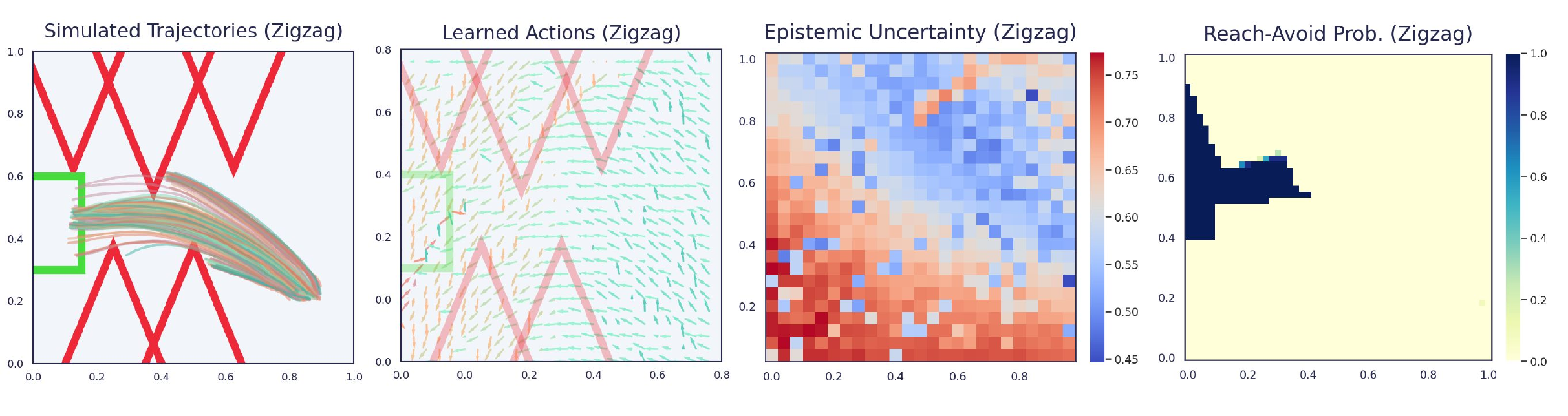}
\caption{\textbf{Left Column:} 200 simulated trajectories for the learned policy starting from the initial state. \textbf{Center Left Column:} A 2D visualization of the learned policy. Each arrow represents the direction of the applied force. \textbf{Center Right Column:} The epistemic uncertainty for the learned dynamics model. \textbf{Right Column:} Certified lower-bounds of probabilistic reach avoid for each abstract state according to BNN and final learned policy.}
\label{fig:LearnedSystems}
\end{figure}

\subsection{Comparing Certification of Learned and Max-Cert Policies}\label{sec:synsthesis_exp}
\begin{figure}
\centering
\includegraphics[width=1.0\textwidth]{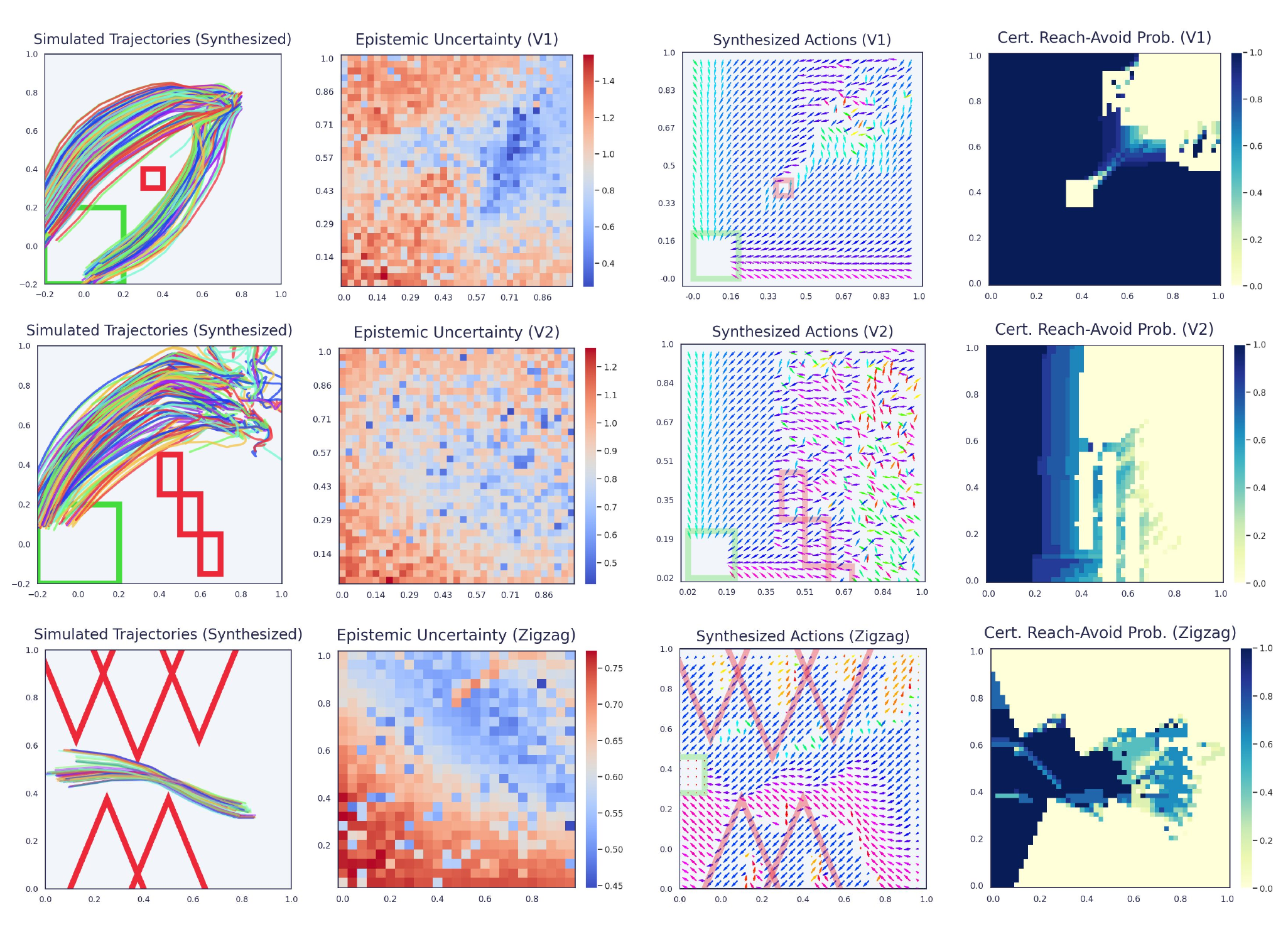}
\caption{
%\np{Caption should limit to describing the legend/color-code and not discuss the results.}\MRW{You are correct. Thank you for the edit, Nicola :)}
A version of each learned system after a new policy has been synthesized from the reach-avoid specification. \textbf{First column:} 200 simulations of the synthesized policy in the real environment. \textbf{Second column:} %\np{pls check}\MRW{Correct, thanks!}
BNN epistemic uncertainty given as the variance of the BNN predictive distribution. \textbf{Third column:} A visualization of the maximally certifiable policies, \sedit{which demonstrate a clearer tendency to avoid obstacles throughout the state space compared to the policies in Figure~\ref{fig:LearnedSystems}.} \textbf{Fourth column:} Synthesized policies have remarkably higher lower-bounds than learned policies, corresponding plots for learned systems in Figure~\ref{fig:LearnedSystems}. %\np{plots for the learned policies are now in appendix.}\ap{I think we should reference in the main text that similar results for the learned policies are in the Appendix.} 
}
\label{fig:SynthesisFigure}
\end{figure}
In Figure~\ref{fig:LearnedSystems} and Figure~\ref{fig:SynthesisFigure} we visualize systems from both learned and synthesized policies. \sedit{Each row represents one of our control environments and is comprised of four figures. These figures show, respectively, simulations from the dynamical system, BNN uncertainty, the control policy plotted as a gradient field, and the certified safety probabilities.} 
%
% \ap{I don't understand why figure 3 is discussed before figure 2. Actually, Figure 2 is never really referred to or discussed properly in the text, but only mentioned in passing.}\MRW{Good point. Notice that these figures make one unified point. Yes most discussion is initiated around Figure 3, but almost always compares and contrasts with Figure 2. I have restructured this section to make this more clear. } 
%\LL{Synthesized how? Using the NN controller or the look up table approach?}
\sedit{The first column of the Figures depicts 200 simulated trajectories of the learned (Figure~\ref{fig:LearnedSystems}) or synthesized (Figure~\ref{fig:SynthesisFigure}) control policies on the BNN (whose uncertainty is plotted along the second column).} Notice how in both cases we visually obtain the behaviour expected, with the overwhelming majority of the simulated trajectories safely avoiding the obstacles (red regions in the figure) and terminating in the goal (green region). A vector field associated with the policy is depicted in the third column \sedit{of the figures}. Notice that, the actions returned by our synthesis method intuitively align with the reach-avoid specification, that is, synthesized actions near the obstacle and out-of-bounds are aligned with moving away from such unsafe states. Exceptions to this are represented by locations where the agent is already unsafe and that, as such, are not fully explored during the BNN learning phase (e.g., the lower triangles in the \texttt{Zigzag} scenario), locations where two directions are equally optimal (e.g., in the top right corner of the \texttt{v1} environment) and locations which are not along any feasibly optimal path (e.g., the lower right corner of \texttt{v2}) and as such are not accounted by the BNN learning.

In Table~\ref{tab:certtable} we compare the certification results of the synthesized policy against the initial learned policy. %\LL{how did you learn it? Give citation to the method used }\np{this should go under exp settings, see my comment above}.
As the synthesized policy is computed by improving on the latter, we expect the former to outperform the learned policy in terms of the guarantees obtained.
%\LL{Next sentence is a bit unclear}\np{better now?}\MRW{Has this been resolved?}
This is in fact confirmed and quantified by the results of Table~\ref{tab:certtable}, which lists, for each of the three environments, \edit{the average reach-avoid probability \sedit{estimated} over 500 trajectories, the average certification lower bound across the state space, and the certification coverage (i.e., the proportion of states where our algorithm returns a non-zero probability lower bound).}  \sedit{This notion of coverage only requires a state to be certified with a probability above 0, and so it is most informative when evaluated together with the average lower-bound and visual inspection of Figure~\ref{fig:LearnedSystems} and Figure~\ref{fig:SynthesisFigure}. }
% the statistical performance of the model/policy obtained (computed over 500 trajectories), the average certification lower bound obtained across the state space, and the certification coverage, across the three environments studied. 

Indeed, the synthesized policy significantly improves on the certification guarantees given by the learned policy, and consistently so across the three environments analysed, with the lower bound improving by a factor of roughly $3.5$. \sedit{This considerable improvement is to be expected as worst-case guarantees can be poor for deep learning systems that are not trained with specific safety objectives}\citep{mirman2018differentiable, gowal2018effectiveness, wicker2021bayesian}. \sedit{In particular, for both the \texttt{V1} and \texttt{Zigzag} case studies, we observe that the average lower bound jumps from roughly 0.2 to greater than 0.7. Moreover, } the most significant improvements are obtained in the most challenging case, i.e., the \texttt{Zigzag} environment, with the certification coverage increasing of a $4.75$ factor.
Interestingly, also the average model performance increases for the synthesized models. Intuitively this occurs because while in the learning of the initial policy passing through the obstacle is only penalised by a continuous factor, the synthesized policy strives to rigorously enforce safety across the BNN posterior. \edit{A visual representation of these results is provided in the last column of Figure~\ref{fig:LearnedSystems} for the learned policy and in Figure~\ref{fig:SynthesisFigure} for the synthesized policy. We note that the uncertainty maps in these figures are identical as the BNN model is not changed, only the policy.} % \ap{This is repetitive from everything that was explained above.} We find that the results on the learned systems are noticeably worse than those for optimal policies, but this is to be expected as worst-case guarantees can be poor for naturally trained deep learning systems \citep{mirman2018differentiable, gowal2018effectiveness, wicker2021bayesian}.}

\begin{table}
\centering
\begin{tabular}{lccc}
\multicolumn{4}{c}{\textit{\textbf{Learned Policy}}}                                                                                                           \\ \cline{2-4} 
\multicolumn{1}{l|}{}                   & \multicolumn{1}{c|}{Performance}     & \multicolumn{1}{c|}{Avg. Lower Bound} & \multicolumn{1}{c|}{Cert. Coverage}         \\ \hline
\multicolumn{1}{|l|}{$\texttt{V1}$}     & \multicolumn{1}{c|}{0.789}           & \multicolumn{1}{c|}{0.212}            & \multicolumn{1}{c|}{0.639}            \\ \hline
\multicolumn{1}{|l|}{$\texttt{V2}$}     & \multicolumn{1}{c|}{0.805}           & \multicolumn{1}{c|}{0.192}            & \multicolumn{1}{c|}{0.484}            \\ \hline
\multicolumn{1}{|l|}{$\texttt{Zigzag}$} & \multicolumn{1}{c|}{0.815}           & \multicolumn{1}{c|}{0.189}            & \multicolumn{1}{c|}{0.193}            \\ \hline
                                        & \multicolumn{1}{l}{}                 & \multicolumn{1}{l}{}                  & \multicolumn{1}{l}{}                  \\
\multicolumn{4}{c}{\textit{\textbf{Synthesized Policy (Optimal)}}}                                                                                             \\ \cline{2-4} 
\multicolumn{1}{l|}{}                   & \multicolumn{1}{c|}{Performance}     & \multicolumn{1}{c|}{Avg. Lower Bound} & \multicolumn{1}{c|}{Cert. Coverage}         \\ \hline
\multicolumn{1}{|l|}{$\texttt{V1}$}     & \multicolumn{1}{c|}{$\mathbf{0.94}$} & \multicolumn{1}{c|}{$\mathbf{0.789}$} & \multicolumn{1}{c|}{$\mathbf{0.808}$} \\ \hline
\multicolumn{1}{|l|}{$\texttt{V2}$}     & \multicolumn{1}{c|}{$\mathbf{0.94}$} & \multicolumn{1}{c|}{$\mathbf{0.597}$} & \multicolumn{1}{c|}{$\mathbf{0.624}$} \\ \hline
\multicolumn{1}{|l|}{$\texttt{Zigzag}$} & \multicolumn{1}{c|}{$\mathbf{1.00}$} & \multicolumn{1}{c|}{$\mathbf{0.710}$} & \multicolumn{1}{c|}{$\mathbf{0.910}$} \\ \hline
\end{tabular}
\caption{\label{tab:certtable} Certification comparisons for learned and synthesized policy across the three environments. \textbf{Performance} indicates the proportion of simulated trajectory that respect the reach-avoid specification. \textbf{Avg.\ Lower Bound} is the mean certification probability across all states. \textbf{Cert.\ Coverage} is the proportion of states that we are able to certify (i.e., with a non-zero lower bound for the reach-avoid probability).}
\end{table}

\begin{table}[]
\centering
\begin{tabular}{lccc}
\multicolumn{4}{c}{\textit{\textbf{Learned Policy}}}                                                                                                                 \\ \cline{2-4} 
\multicolumn{1}{l|}{}                         & \multicolumn{1}{c|}{Performance}     & \multicolumn{1}{c|}{Avg. Lower Bound} & \multicolumn{1}{c|}{Coverage}         \\ \hline
\multicolumn{1}{|l|}{Var. Inference}   & \multicolumn{1}{c|}{0.832}           & \multicolumn{1}{c|}{0.399}            & \multicolumn{1}{c|}{0.696}            \\ \hline
\multicolumn{1}{|l|}{Ham. Monte Carlo} & \multicolumn{1}{c|}{0.789}           & \multicolumn{1}{c|}{0.212}            & \multicolumn{1}{c|}{0.639}            \\ \hline
                                              & \multicolumn{1}{l}{}                 & \multicolumn{1}{l}{}                  & \multicolumn{1}{l}{}                  \\
\multicolumn{4}{c}{\textit{\textbf{Synthesized Policy (Optimal)}}}                                                                                                   \\ \cline{2-4} 
\multicolumn{1}{l|}{}                         & \multicolumn{1}{c|}{Performance}     & \multicolumn{1}{c|}{Avg. Lower Bound} & \multicolumn{1}{c|}{Coverage}         \\ \hline
\multicolumn{1}{|l|}{Var. Inference}   & \multicolumn{1}{c|}{$\mathbf{1.00}$} & \multicolumn{1}{c|}{$\mathbf{0.762}$} & \multicolumn{1}{c|}{$\mathbf{0.851}$} \\ \hline
\multicolumn{1}{|l|}{Ham. Monte Carlo} & \multicolumn{1}{c|}{$\mathbf{0.94}$} & \multicolumn{1}{c|}{$\mathbf{0.789}$} & \multicolumn{1}{c|}{$\mathbf{0.808}$} \\ \hline
\end{tabular}
\caption{\label{tab:inftab} Certification comparisons for learned and synthesized policy between VI and HMC BNN learning on obstacle layout \texttt{V1}. \textbf{Performance} indicates the proportion of simulated trajectory that respect the reach-avoid specification. \textbf{Avg.\ Lower Bound} is the mean certification probability across all states. \textbf{Cert.\ Coverage} is the proportion of states that we are able to certify with non-zero probability.}
\end{table}

\begin{figure}[h]
\centering
\includegraphics[width=0.8\textwidth]{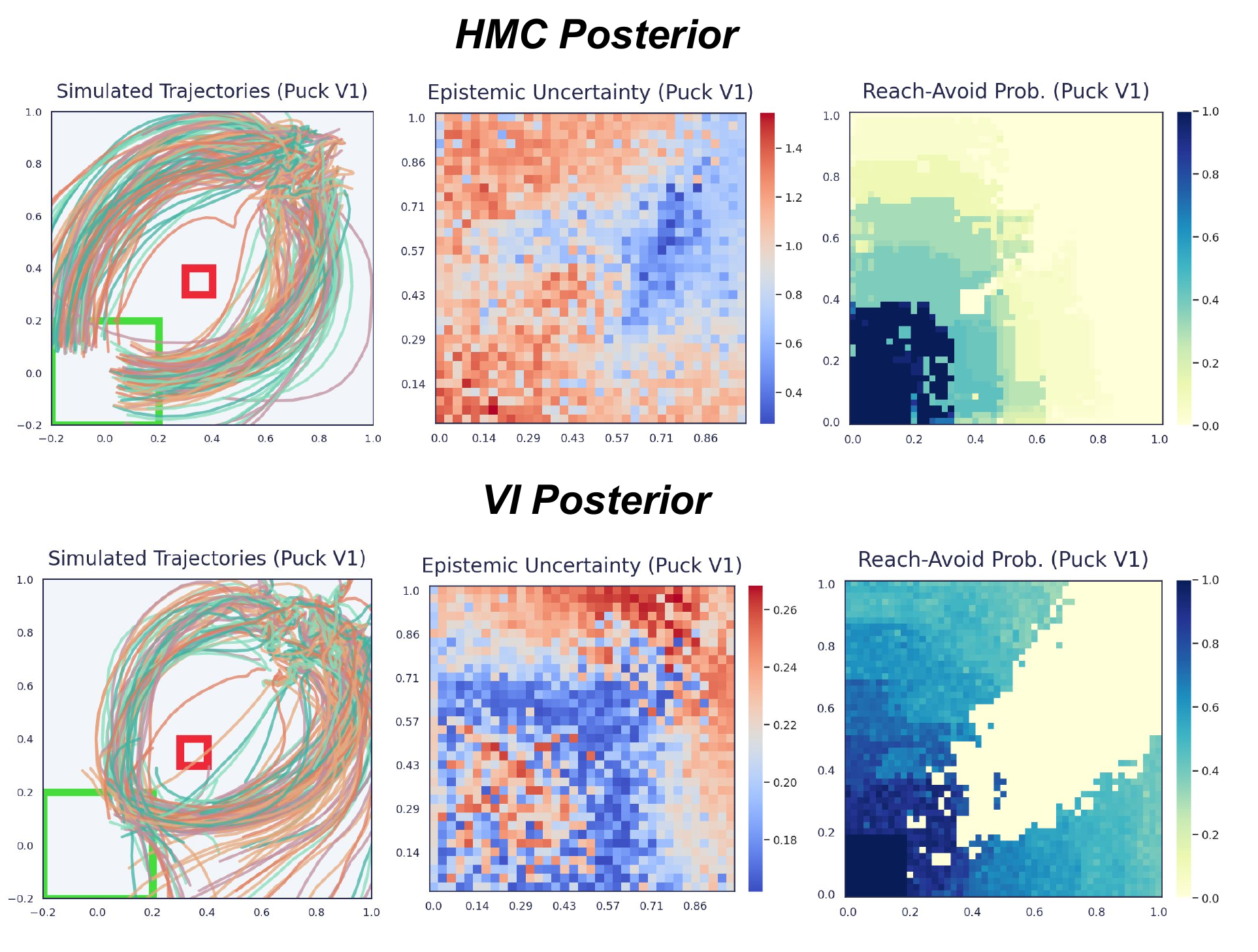}
\caption{\textbf{Top Row:}  Visualization of the learned system using HMC to approximately infer dynamics. \textbf{Bottom Row: } Visualization of the learned system using VI to approximately infer dynamics. We highlight that the VI approximation displays a 5 to 10 times reduction in epistemic uncertainty.%\ap{Why do we highlight this here? I don't fully understand the reason for this comment in the caption of the figure }\MRW{This must be highlighted in text as the color scales between the two figures are not coordinated. Not highlighting this here would be dishonest.}
}\label{fig:ApproxInfLearn}
\end{figure}

\begin{figure}[h]
%\vspace{-5em} %MRW I will need to come back and fix this prior to submission.
\centering
\includegraphics[width=0.8\textwidth]{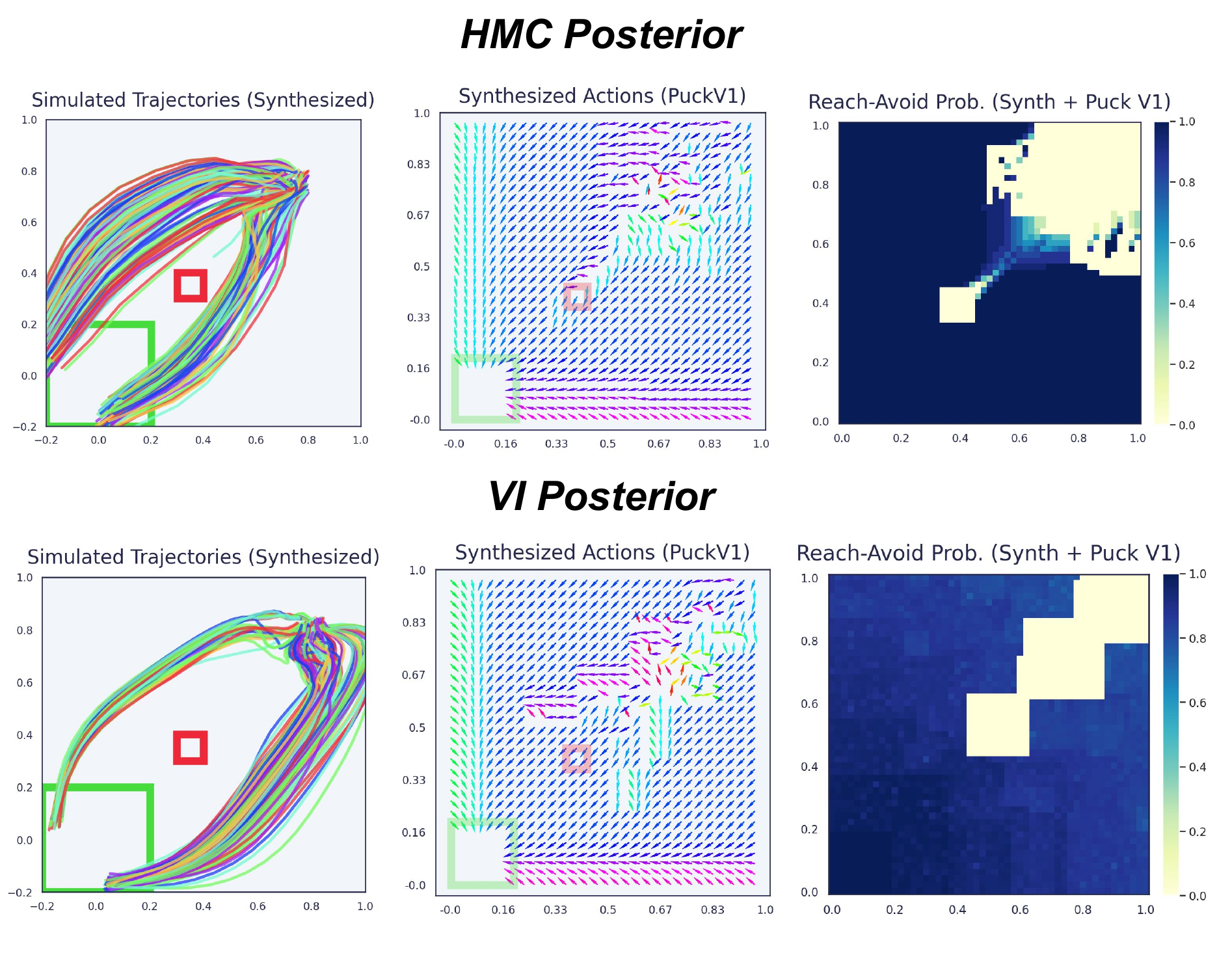}
\caption{\textbf{Top Row:} Visualization of the synthesized policy and its performance based on the HMC dynamics model. \textbf{Bottom Row:} Visualization of the synthesized policy and its performance based on the VI dynamics model.% We highlight that despite better certification in the learned setting, HMC displays stronger guarantees w.r.t.\ a synthesized policy. This is apparent in the greater coverage behind the obstacle and is numerically quantified in Table~\ref{tab:inftab}. \ap{Actually VI has slightly better coverage. HMC has slightly better average lower bound. It doesn't seem to me that one performs better than the other.}
}\label{fig:ApproxInfSynth}
\end{figure}

\subsection{On the Effect of Approximate Inference}\label{sec:res_approx_inf}
%\MK{Would be helpful, but is too sketchy and not systematic}
The results provided so far have been generated with BNN dynamical models learned via HMC training.
However, different inference methods produce different approximations of the BNN posterior thus leading to different dynamics approximations and hence synthesized policies. 

Table~\ref{tab:inftab} and the plots in Figures~\ref{fig:ApproxInfLearn} and \ref{fig:ApproxInfSynth} analyse the effect of approximate inference on both learned and synthesized policies, comparing results obtained by HMC with those obtained by VI training on the \texttt{v1} scenario.
We notice that also in the case of VI the synthesized policy significantly improves on the initial policy.
Interestingly, the certification results over the learned policy for VI are higher than those obtained for HMC, but the results are comparable for the synthesized policies. 
In fact, it is known in the literature that  VI tends to under-estimate uncertainty \citep{myshkov2016posterior, michelmore2020uncertainty} and is more susceptible to model misspecification \citep{masegosa2020learning}. 
As such, being probabilistic, the bound obtained is tighter for VI where the uncertainty is lower than that of HMC which provides a more conservative representation of the agent dynamics.
For example, we see in the first two rows of Table~\ref{tab:inftab} that the average lower bound achieved for the variational inference posterior is 1.88 times higher than the bounds for HMC posterior. 
However, our synthesis method reduces this gap between HMC and VI, while still accounting for the higher uncertainty of the former, and hence the more conservative guarantees. %\np{I thought that VI had smaller uncertainty than HMC, but now we state (and observe) the opposite?}\MRW{Hm, not sure where we would have implied this. I will do a careful read and check.}\np{earlier in this paragraph we write ``the bound obtained is tighter for VI where the uncertainty is lower to that of HMC which provides a more conservative representation''}\ap{Yes I think the erorr is in the very last sentence, where instead of "latter" you should say "former". But have a look}\MRW{Thanks, Andrea, you are right. Thanks Nicola for pointing this out as well}

\begin{figure}[h]
\centering
\includegraphics[width=0.9\textwidth]{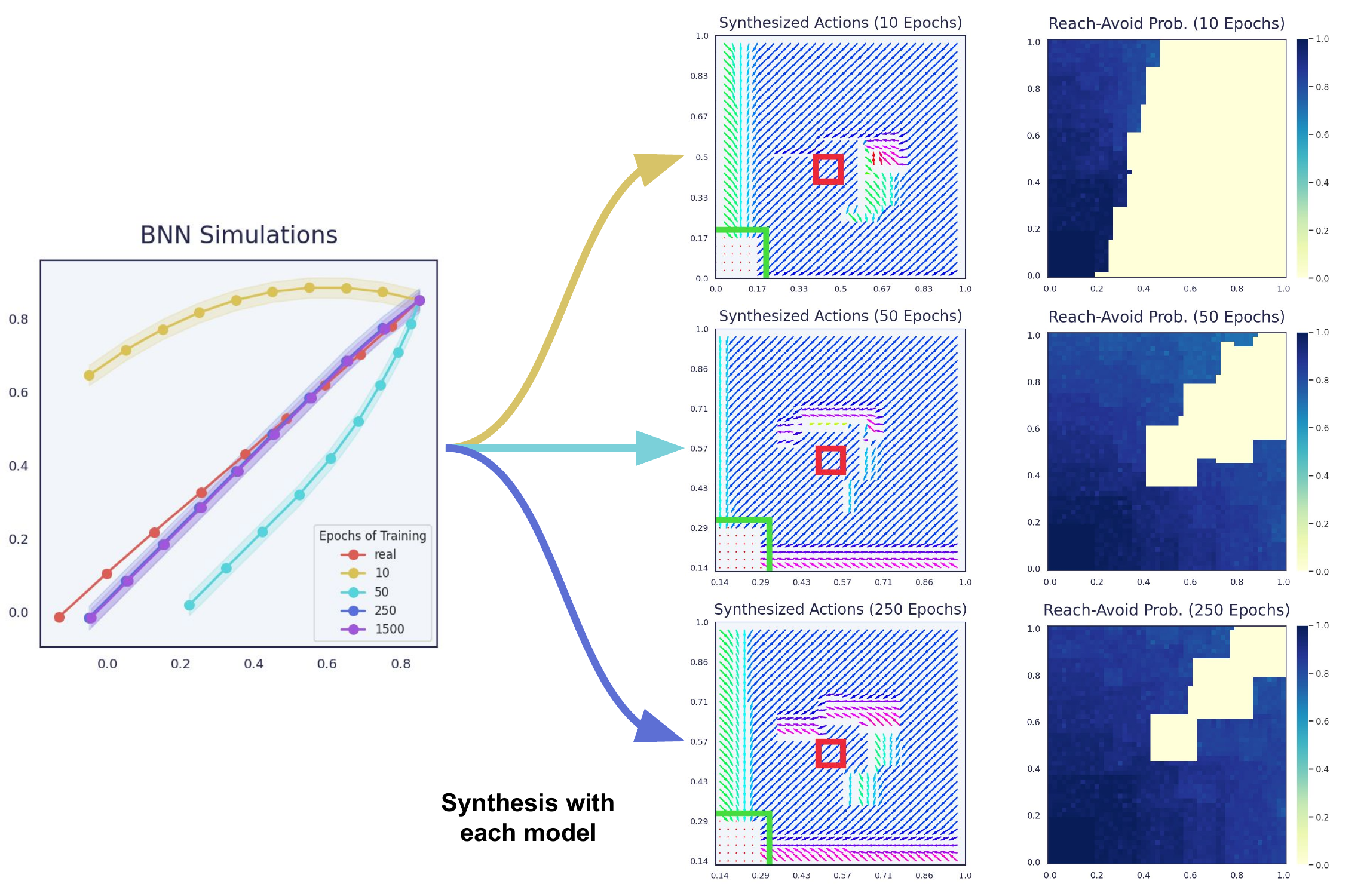}
\caption{Analysis for number of training epochs used in performing VI training on the \texttt{V1} environment. \edit{Left: sample of 10-step agent trajectories obtained with BNNs trained with VI and different number of epochs (red: ground truth trajectory). Right: synthesis and certification results for a selection of training epochs.}}
\label{fig:ModelCalibration}
\end{figure}

While HMC approximates the posterior by relying on a Monte Carlo estimate of it, VI is a gradient-based technique, where the number of training epochs (i.e., the number of full sweeps through the dataset) is a key hyper-parameter. We thus analyse the effect of training epochs in the quality of the dynamics obtained in Figure~\ref{fig:ModelCalibration}, along with the effect on the synthesized policies and the certificates obtained for such policies.  
The left plot of the figure shows a set of predicted trajectories over a 10 time-step horizon for a varying number of training epochs, with the ground truth behaviour highlighted in red. The BNN trajectories are color-coded based on the number of epochs each dynamics model was trained for. In yellow, we see that the BNN which has only been trained for 10 epochs displays considerable error in its iterative predictions. This is reduced considerably for a model trained for 50 epochs, but the cumulative error after 10 epochs is still considerable. Finally, as expected, for models trained for 250 and 1500 epochs we empirically observe a trend toward convergence to the ground truth. 

We notice that the policy and certifications directly reflect the quality of the approximation. In fact, as we increase the model fit, we see that there are significant improvements in both the intuitive behavior of the synthesized policy as well as the resulting guarantees we are able to compute.

\begin{figure}[h]
\centering
\includegraphics[width=0.95\textwidth]{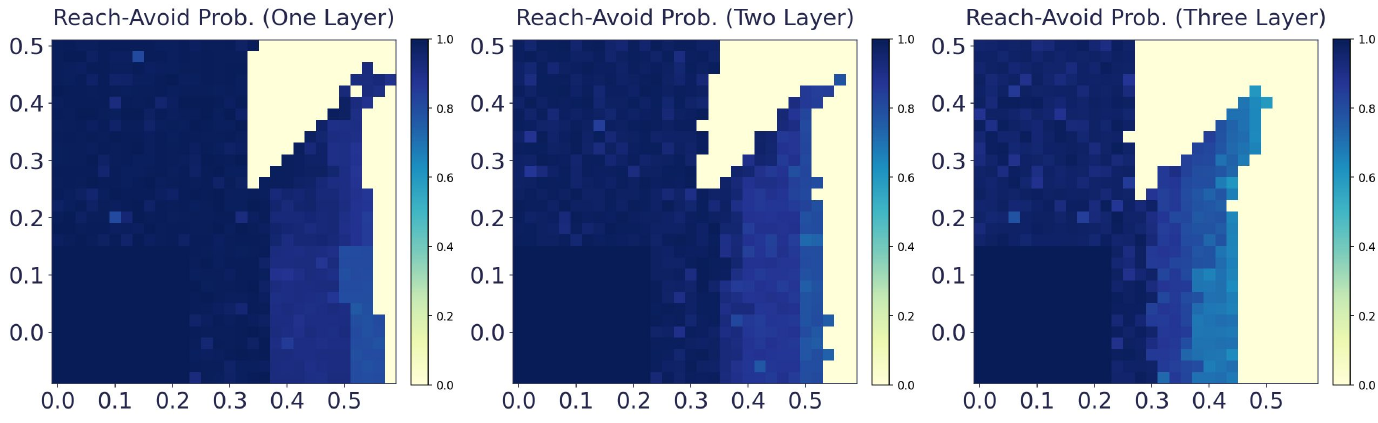}
\caption{\edit{We vary the depth of the BNN used to learn the dynamics and observe its effects on our certified safety probabilities over the first half of the Puck-V1 state-space. From left to right we plot the lower-bound reach-avoid probabilities for a one-layer BNN dynamics model, a two-layer BNN dynamics model, and a three-layer BNN dynamics model. }}
\label{fig:DepthExperiments}
\end{figure}

\subsection{Depth Experiments}\label{sec:scalability}
\edit{In this section, we evaluate how our method performs when we vary the depth of the BNN dynamics model considered. In Figure~\ref{fig:DepthExperiments}, we plot the certified reach-avoid probabilities for a learned policy and a one, two, and three-layer BNN dynamics model where each layer has a width of 12 neurons. Similar architectures are found in the BNNs studied in recent related work \citep{lechner2021infinite}. Other than the depth of the BNN, all the other variables in the experiment are held equal (e.g., number of episodes during learning, discretization of state-space, and number of BNN samples considered for the lower bound). 
%\ap{You have to say here how many neuorns you use. Btw calling three layers with 12 neurons scalability analysis is a bit of a stretch – if I were a reviewer I'd go absolutely mad for something like that. We have two possibility in front of us: either (1) you provide a good amount of citations which show that these are the architectrues used in practice (but I doubt that they are) or (2) We call this section something else, like analysis of the effect of depth on certification, and we make sure not to refer to it as scalability analysis.}
The learning procedure results in BNN models with roughly equivalent losses and in policies that are qualitatively similar, see  \ref{app:scalability} for further visualizations. Given that the key factors of the system have been held equal, we notice a decrease in our certified lower bound as depth increases. Specifically, the average lower bound for the one-layer model is 0.811, for the two-layer model it is 0.763, and for the three-layer model it decreases further to 0.621. Figure~\ref{fig:DepthExperiments} clearly demonstrates that as the BNN dynamics model becomes deeper, then our lower bound becomes more conservative. This finding is consistent with existing results in certification of BNNs \citep{wicker2020probabilistic, berrada2021make} and DNNs \citep{gowal2018effectiveness,mirman2018differentiable}. We note, however, that when the verification parameters are refined, i.e., more samples from the BNN are taken, we are able to certify the three-layer system with an average certified lower-bound of 0.867 (see \ref{app:scalability}). These additional BNN samples increase the runtime of our certification procedure by 1.5 times.  %\ap{This is clearly not true, especially since it is a network with 12 neurons per layer. You can't take big conclusion out of this. Rephrase it in a more conservative way. With depth clearly the problems gets more complicated, but the bound can still be refined better with more time, however this becomes computationally infeasible very quickly.}
}

\section{Related Work}\label{sec:related}

%\MK{Needs restructuring into thematic groupings and improving by more citations, rather sketchy and focused on own work}

%\LL{Nicola, can you add the papers you were talking about(especially in the last paragraph)? I already added some and revised also following Marta's comments.}

Certification of machine learning models is a rapidly growing area~\citep{gehr2018ai2,katz2017reluplex,gowal2018effectiveness, wicker2022robust}. %, underpinned by techniques ranging from SMT solving, to abstract interpretation, to convex optimisation, to testing, and to constraint propagation and relaxation~\citep{ehlers2017formal,huang2017safety,katz2017reluplex,katz2019marabou,elboher2020abstraction,wang2018efficient,gehr2018ai2,singh2019abstract,wang2018formal,bak2020improved,DBLP:conf/cav/TranBXJ20}. %\np{other work by Marta's or other groups, on NN verification?} 
While most of these methods have been designed for deterministic NNs,
recently safety analysis of Bayesian machine learning models has been studied both for Gaussian processes (GPs)~\citep{grosse2017wrong,cardelli2018robustness,blaas2020adversarial} and BNNs~\citep{athalye2018obfuscated,cardelli2019statistical, wicker2020probabilistic},  including methods for adversarial training  \citep{liu2018adv,wicker2021bayesian}. %In particular, \citep{wicker2020probabilistic} uses constraint propagation techniques~\citep{gowal2018effectiveness,weng2018towards,zhang2018efficient} to derive a guaranteed lower bound on the probability that all the points in an input set are mapped by the BNN to a given output region, a result that we shall leverage in the present work (see Section \ref{sec:LowerBound}). 
%longer text
% The first solution method for probabilistic safety of BNNs was recently presented in~\citep{wicker2020probabilistic}. This technique (used in the present work to compute one-step probabilities, see Prop.~\ref{Proposition:UAI}) builds on constraint propagation over NNs~\citep{gowal2018effectiveness,weng2018towards,zhang2018efficient} and derives a guaranteed lower bound on the probability that the neural network prediction remains in a given output region for any point in a given input region. 
%\np{other works on formal analysis of Bayesian models?}\acmt{not that I'm aware of.}
The above works, however, focus exclusively on the input-output behaviour of the models, that is, can only reason about static properties. 
Conversely, the problem we tackle in this work has additional complexity, as we aim to formally reason about iterative predictions, i.e., trajectory-level behaviour of a BNN interacting in a closed loop with a controller.

Iterative predictions have been widely studied for Gaussian processes  \citep{girard2003gaussian} and safety guarantees have been proposed in this setting in the context of model-based RL with GPs \citep{jackson2020safety,polymenakos2019safe,berkenkamp2017safe,berkQuad}. %\LL{Add berkenkamp2017 in the cover letter}\MRW{Done :) }
%Of particular interest for our paper is the work in \cite{polymenakos2019safe} that employs Gaussian Processes to model control systems, and a model-based RL architecture to synthesise safe and optimal policies, which are further attached with guarantees in \cite{polymenakos2020safety}.
However, all these works are specific to GPs and cannot be extended to BNNs, whose posterior predictive distribution is intractable and non-Gaussian even for the more commonly employed approximate Bayesian inference methods \citep{neal2012bayesian}. Recently, iterative prediction of \emph{neural network dynamic models} have been studied \citep{wei2021safe,adams2022formal} and methods to certify these models against temporal logic formulae have been derived \citep{adams2022formal}.
However, these works only focus on standard (i.e., non-Bayesian) neural networks with additive Gaussian noise. Closed-loop systems with known (deterministic) models and control policies modelled as BNNs are considered in \citep{lechner2021infinite}. In contrast with our work,  \cite{lechner2021infinite}  can only support deterministic models without noisy dynamics, only focus on the safety verification problem, and are limited to BNN posterior with unimodal weight distribution.

Various recent works consider verification or synthesis of RL schemes against reachability specifications 
% with deterministic  NNs in the loop 
\citep{sun2019formal,konighofer2020safe,bacci2020probabilistic}.
%For instance, safe RL methods have been proposed that monitor at runtime the actions of a given RL agent and correct them (with a safe action or by switching to a failsafe agent) whenever such actions can lead to violating the specification~\citep{alshiekh2018safe,avni2019run,bouton2019reinforcement,phan2020neural}.
None of these approaches, however, support both continuous state-action spaces and probabilistic models, as in this work. 
Continuous action spaces are supported in \citep{hasanbeig2020certified}, where the authors provide RL schemes for the synthesis of policies maximising given temporal requirements, which is also extended to continuous state- and action-spaces in \citep{bcHKA20}. However, the guarantees resulting from these model-free algorithms are asymptotic and thus of a different nature than those in this work.  
%Related work concerns reachability analysis of learning-enabled control models~\citep{DBLP:conf/cav/TranYLMNXBJ20,huang2019reachnn,ivanov2019verisig,dutta2019reachability,sun2019formal}, i.e., systems consisting of a deterministic NN control policy and a given ODE-based environment model: this work generalises these results in computing probabilistic reachability for systems with NN-based policies and data-learnt BNN models, thus formally accounting for learning uncertainty;  it also synthesises correct-by-design NN policies for said systems.  
The work of \cite{HVA17} integrates Bayesian inference and formal verification over control models, additionally proposing strategy synthesis approaches for active learning~\citep{bcWA19}. In contrast to our paper these works do not support unknown noisy models learned via BNNs. 

A related line of work concerns the synthesis of runtime monitors for predicting the safety of the policy's actions and, if necessary, correct them with fail-safe actions \citep{alshiekh2018safe,avni2019run,bouton2019reinforcement,fulton2019verifiably,phan2020neural}. These approaches, however, do not support continuous state-action spaces or require some form of ground-truth mechanistic model for safety verification (as opposed to our data-driven BNN models).

\section{Conclusions}
%\MK{Needs to be more concerete, revise}
In this paper, we considered the problem of computing the probability of time-bounded reach-avoid specifications for dynamic models described by iterative predictions of BNNs. 
We developed methods and algorithms to
compute a lower bound of this reach-avoid probability. Additionally, relying on techniques
from dynamic programming and non-convex optimization, we 
synthesized certified controllers that maximize probabilistic reach-avoid.
In a set of experiments, we showed that our framework enables
certification of strategies on non-trivial control tasks. \sedit{A future research direction will be to investigate techniques to enhance the scalability of our methods so that they can be applied to state-of-the-art reinforcement learning environments. However, we emphasise that the benchmark considered in this work remains a challenging one for certification purposes, due to both the non-linearity and stochasticity of the models, and the sequential, multi-step dependency of the predictions.} Thus, this paper makes an important step toward the application of BNNs in safety-critical scenarios. 
%\np{I removed sentence on future work}\MRW{Thanks!}
% In future, we plan to consider more complex properties expressed in temporal logic specification such as Linear Temporal Logic \citep{baier2008principles}.

\section*{Acknowledgements}
%\MK{Use ack command or include at the end}
This project received funding from the ERC under the European Union’s Horizon 2020 research and innovation programme (FUN2MODEL, grant agreement No.~834115).

%\appendix



%\bibliographystyle{plainnat} % We choose the "plain" reference style
\bibliography{cite} % Entries are in the refs.bib file

\begin{thebibliography}{67}
\providecommand{\natexlab}[1]{#1}
\providecommand{\url}[1]{\texttt{#1}}
\expandafter\ifx\csname urlstyle\endcsname\relax
  \providecommand{\doi}[1]{doi: #1}\else
  \providecommand{\doi}{doi: \begingroup \urlstyle{rm}\Url}\fi

\bibitem[Abate et~al.(2008)Abate, Prandini, Lygeros, and Sastry]{abate2008probabilistic}
Alessandro Abate, Maria Prandini, John Lygeros, and Shankar Sastry.
\newblock Probabilistic reachability and safety for controlled discrete time stochastic hybrid systems.
\newblock \emph{Automatica}, 44\penalty0 (11):\penalty0 2724--2734, 2008.

\bibitem[Adams et~al.(2022)Adams, Lahijanian, and Laurenti]{adams2022formal}
Steven Adams, Morteza Lahijanian, and Luca Laurenti.
\newblock Formal control synthesis for stochastic neural network dynamic models.
\newblock \emph{arXiv preprint arXiv:2203.05903}, 2022.

\bibitem[Alshiekh et~al.(2018)Alshiekh, Bloem, Ehlers, K{\"o}nighofer, Niekum, and Topcu]{alshiekh2018safe}
Mohammed Alshiekh, Roderick Bloem, R{\"u}diger Ehlers, Bettina K{\"o}nighofer, Scott Niekum, and Ufuk Topcu.
\newblock Safe reinforcement learning via shielding.
\newblock In \emph{Proceedings of the AAAI Conference on Artificial Intelligence}, volume~32, 2018.

\bibitem[Astrom and Murray(2008)]{Astrom08}
Karl~J. Astrom and Richard~M. Murray.
\newblock \emph{Feedback Systems: An Introduction for Scientists and Engineers}.
\newblock Princeton University Press, Princeton, NJ, USA, 2008.

\bibitem[Athalye et~al.(2018)Athalye, Carlini, and Wagner]{athalye2018obfuscated}
Anish Athalye, Nicholas Carlini, and David Wagner.
\newblock Obfuscated gradients give a false sense of security: Circumventing defenses to adversarial examples.
\newblock In \emph{International Conference on Machine Learning}, pages 274--283. PMLR, 2018.

\bibitem[Avni et~al.(2019)Avni, Bloem, Chatterjee, Henzinger, K{\"o}nighofer, and Pranger]{avni2019run}
Guy Avni, Roderick Bloem, Krishnendu Chatterjee, Thomas~A Henzinger, Bettina K{\"o}nighofer, and Stefan Pranger.
\newblock Run-time optimization for learned controllers through quantitative games.
\newblock In \emph{International Conference on Computer Aided Verification}, pages 630--649. Springer, 2019.

\bibitem[Bacci and Parker(2020)]{bacci2020probabilistic}
Edoardo Bacci and David Parker.
\newblock Probabilistic guarantees for safe deep reinforcement learning.
\newblock In \emph{International Conference on Formal Modeling and Analysis of Timed Systems}, pages 231--248. Springer, 2020.

\bibitem[Berkenkamp et~al.(2016)Berkenkamp, Schoellig, and Krause]{berkQuad}
Felix Berkenkamp, Angela~P. Schoellig, and Andreas Krause.
\newblock Safe controller optimization for quadrotors with {G}aussian processes.
\newblock In \emph{Proc. of the IEEE International Conference on Robotics and Automation (ICRA)}, pages 493--496, 2016.

\bibitem[Berkenkamp et~al.(2017)Berkenkamp, Turchetta, Schoellig, and Krause]{berkenkamp2017safe}
Felix Berkenkamp, Matteo Turchetta, Angela~P Schoellig, and Andreas Krause.
\newblock Safe model-based reinforcement learning with stability guarantees.
\newblock In \emph{NIPS}, 2017.

\bibitem[Berrada et~al.(2021)Berrada, Dathathri, Dvijotham, Stanforth, Bunel, Uesato, Gowal, and Kumar]{berrada2021make}
Leonard Berrada, Sumanth Dathathri, Krishnamurthy Dvijotham, Robert Stanforth, Rudy~R Bunel, Jonathan Uesato, Sven Gowal, and M~Pawan Kumar.
\newblock Make sure you're unsure: A framework for verifying probabilistic specifications.
\newblock \emph{Advances in Neural Information Processing Systems}, 34:\penalty0 11136--11147, 2021.

\bibitem[Bertsekas and Shreve(2004)]{bertsekas2004stochastic}
Dimitir~P Bertsekas and Steven Shreve.
\newblock \emph{Stochastic optimal control: the discrete-time case}.
\newblock Athena Scientific, 2004.

\bibitem[Blaas et~al.(2020)Blaas, Patane, Laurenti, Cardelli, Kwiatkowska, and Roberts]{blaas2020adversarial}
Arno Blaas, Andrea Patane, Luca Laurenti, Luca Cardelli, Marta Kwiatkowska, and Stephen Roberts.
\newblock Adversarial robustness guarantees for classification with {G}aussian processes.
\newblock In \emph{International Conference on Artificial Intelligence and Statistics}, pages 3372--3382. PMLR, 2020.

\bibitem[Blundell et~al.(2015)Blundell, Cornebise, Kavukcuoglu, and Wierstra]{blundell2015weight}
Charles Blundell, Julien Cornebise, Koray Kavukcuoglu, and Daan Wierstra.
\newblock Weight uncertainty in neural network.
\newblock In \emph{International Conference on Machine Learning}, pages 1613--1622. PMLR, 2015.

\bibitem[Bouton et~al.(2019)Bouton, Karlsson, Nakhaei, Fujimura, Kochenderfer, and Tumova]{bouton2019reinforcement}
Maxime Bouton, Jesper Karlsson, Alireza Nakhaei, Kikuo Fujimura, Mykel~J Kochenderfer, and Jana Tumova.
\newblock Reinforcement learning with probabilistic guarantees for autonomous driving.
\newblock \emph{arXiv preprint arXiv:1904.07189}, 2019.

\bibitem[Carbone et~al.(2020)Carbone, Wicker, Laurenti, Patane\textquotesingle, Bortolussi, and Sanguinetti]{carbone2020robustness}
Ginevra Carbone, Matthew Wicker, Luca Laurenti, Andrea Patane\textquotesingle, Luca Bortolussi, and Guido Sanguinetti.
\newblock Robustness of bayesian neural networks to gradient-based attacks.
\newblock In H.~Larochelle, M.~Ranzato, R.~Hadsell, M.~F. Balcan, and H.~Lin, editors, \emph{Advances in Neural Information Processing Systems}, volume~33, pages 15602--15613. Curran Associates, Inc., 2020.

\bibitem[Cardelli et~al.(2019{\natexlab{a}})Cardelli, Kwiatkowska, Laurenti, Paoletti, Patane, and Wicker]{cardelli2019statistical}
Luca Cardelli, Marta Kwiatkowska, Luca Laurenti, Nicola Paoletti, Andrea Patane, and Matthew Wicker.
\newblock Statistical guarantees for the robustness of bayesian neural networks.
\newblock In \emph{Proceedings of the Twenty-Eighth International Joint Conference on Artificial Intelligence, {IJCAI-19}}, pages 5693--5700. International Joint Conferences on Artificial Intelligence Organization, 7 2019{\natexlab{a}}.
\newblock \doi{10.24963/ijcai.2019/789}.
\newblock URL \url{https://doi.org/10.24963/ijcai.2019/789}.

\bibitem[Cardelli et~al.(2019{\natexlab{b}})Cardelli, Kwiatkowska, Laurenti, and Patane]{cardelli2018robustness}
Luca Cardelli, Marta Kwiatkowska, Luca Laurenti, and Andrea Patane.
\newblock Robustness guarantees for {B}ayesian inference with {G}aussian processes.
\newblock In \emph{Proceedings of the AAAI Conference on Artificial Intelligence}, volume~33, pages 7759--7768, 2019{\natexlab{b}}.

\bibitem[Cauchi et~al.(2019)Cauchi, Laurenti, Lahijanian, Abate, Kwiatkowska, and Cardelli]{cauchi2019efficiency}
Nathalie Cauchi, Luca Laurenti, Morteza Lahijanian, Alessandro Abate, Marta Kwiatkowska, and Luca Cardelli.
\newblock Efficiency through uncertainty: Scalable formal synthesis for stochastic hybrid systems.
\newblock In \emph{Proceedings of the 22nd ACM International Conference on Hybrid Systems: Computation and Control}, pages 240--251, 2019.

\bibitem[Deisenroth and Rasmussen(2011)]{pilco}
Marc~Peter Deisenroth and Carl~Edward Rasmussen.
\newblock {PILCO}: {A} model-based and data-efficient approach to policy search.
\newblock In \emph{In Proceedings of the International Conference on Machine Learning}, 2011.

\bibitem[Depeweg et~al.(2017)Depeweg, Hern{\'a}ndez-Lobato, Doshi-Velez, and Udluft]{depeweg2016learning}
S~Depeweg, JM~Hern{\'a}ndez-Lobato, F~Doshi-Velez, and S~Udluft.
\newblock Learning and policy search in stochastic dynamical systems with bayesian neural networks.
\newblock In \emph{5th International Conference on Learning Representations, ICLR 2017-Conference Track Proceedings}, 2017.

\bibitem[Fubini(1907)]{fubini1907sugli}
Guido Fubini.
\newblock Sugli integrali multipli.
\newblock \emph{Rend. Acc. Naz. Lincei}, 16:\penalty0 608--614, 1907.

\bibitem[Fulton and Platzer(2019)]{fulton2019verifiably}
Nathan Fulton and Andr{\'e} Platzer.
\newblock Verifiably safe off-model reinforcement learning.
\newblock In \emph{International Conference on Tools and Algorithms for the Construction and Analysis of Systems}, pages 413--430. Springer, 2019.

\bibitem[Gal et~al.(2016{\natexlab{a}})Gal, McAllister, and Rasmussen]{gal2016improving}
Yarin Gal, Rowan McAllister, and Carl~Edward Rasmussen.
\newblock Improving {PILCO} with {B}ayesian neural network dynamics models.
\newblock In \emph{International Conference in Machine Learning {(ICML)}}, 2016{\natexlab{a}}.

\bibitem[Gal et~al.(2016{\natexlab{b}})Gal, McAllister, and Rasmussen]{pilconn}
Yarin Gal, Rowan~Thomas McAllister, and Carl~Edward Rasmussen.
\newblock Improving {PILCO} with {Bayesian} neural network dynamics models.
\newblock In \emph{Data-Efficient Machine Learning workshop}, volume 951, page 2016, 2016{\natexlab{b}}.

\bibitem[Gehr et~al.(2018)Gehr, Mirman, Drachsler-Cohen, Tsankov, Chaudhuri, and Vechev]{gehr2018ai2}
Timon Gehr, Matthew Mirman, Dana Drachsler-Cohen, Petar Tsankov, Swarat Chaudhuri, and Martin Vechev.
\newblock Ai2: Safety and robustness certification of neural networks with abstract interpretation.
\newblock In \emph{2018 IEEE S\&P}, pages 3--18. IEEE, 2018.

\bibitem[Girard et~al.(2003)Girard, Rasmussen, Candela, and Murray-Smith]{girard2003gaussian}
Agathe Girard, Carl~Edward Rasmussen, Joaquin~Quinonero Candela, and Roderick Murray-Smith.
\newblock {G}aussian process priors with uncertain inputs application to multiple-step ahead time series forecasting.
\newblock In \emph{Advances in neural information processing systems}, pages 545--552, 2003.

\bibitem[Glorot and Bengio(2010)]{glorot2010understanding}
Xavier Glorot and Yoshua Bengio.
\newblock Understanding the difficulty of training deep feedforward neural networks.
\newblock In \emph{Proceedings of the thirteenth international conference on artificial intelligence and statistics}, pages 249--256. JMLR Workshop and Conference Proceedings, 2010.

\bibitem[Goodfellow et~al.(2016)Goodfellow, Bengio, and Courville]{goodfellow2016deep}
Ian Goodfellow, Yoshua Bengio, and Aaron Courville.
\newblock \emph{Deep learning}.
\newblock MIT press, 2016.

\bibitem[Gowal et~al.(2018)Gowal, Dvijotham, Stanforth, Bunel, Qin, Uesato, Arandjelovic, Mann, and Kohli]{gowal2018effectiveness}
Sven Gowal, Krishnamurthy Dvijotham, Robert Stanforth, Rudy Bunel, Chongli Qin, Jonathan Uesato, Relja Arandjelovic, Timothy Mann, and Pushmeet Kohli.
\newblock On the effectiveness of interval bound propagation for training verifiably robust models.
\newblock \emph{Neural Information Processing Systems (NeurIPS)}, 2018.

\bibitem[Grosse et~al.(2017)Grosse, Pfaff, Smith, and Backes]{grosse2017wrong}
Kathrin Grosse, David Pfaff, Michael~Thomas Smith, and Michael Backes.
\newblock How wrong am i?-studying adversarial examples and their impact on uncertainty in gaussian process machine learning models.
\newblock \emph{arXiv preprint arXiv:1711.06598}, 2017.

\bibitem[Haesaert et~al.(2017)Haesaert, Hof, and Abate]{HVA17}
S.~Haesaert, P.M.J.~V.d. Hof, and A.~Abate.
\newblock Data-driven and model-based verification via {B}ayesian identification and reachability analysis.
\newblock \emph{Automatica}, 79\penalty0 (5):\penalty0 115--126, 2017.

\bibitem[Hasanbeig et~al.(2020)Hasanbeig, Kroening, and Abate]{bcHKA20}
M.~Hasanbeig, D.~Kroening, and A.~Abate.
\newblock Deep reinforcement learning with temporal logics.
\newblock In \emph{Proceedings of FORMATS, LNCS 12288}, pages 1--22, 2020.

\bibitem[Hasanbeig et~al.(2019)Hasanbeig, Abate, and Kroening]{hasanbeig2020certified}
Mohammadhosein Hasanbeig, Alessandro Abate, and Daniel Kroening.
\newblock Certified reinforcement learning with logic guidance.
\newblock \emph{arXiv preprint arXiv:1902.00778}, 2019.

\bibitem[Huang and Rosendo(2020)]{huang2020deep}
Jingyi Huang and Andre Rosendo.
\newblock Deep vs. deep bayesian: Reinforcement learning on a multi-robot competitive experiment.
\newblock \emph{arXiv preprint arXiv:2007.10675}, 2020.

\bibitem[Izmailov et~al.(2021)Izmailov, Vikram, Hoffman, and Wilson]{izmailov2021bayesian}
Pavel Izmailov, Sharad Vikram, Matthew~D Hoffman, and Andrew Gordon~Gordon Wilson.
\newblock What are bayesian neural network posteriors really like?
\newblock In \emph{International conference on machine learning}, pages 4629--4640. PMLR, 2021.

\bibitem[Jackson et~al.(2020)Jackson, Laurenti, Frew, and Lahijanian]{jackson2020safety}
John Jackson, Luca Laurenti, Eric Frew, and Morteza Lahijanian.
\newblock Safety verification of unknown dynamical systems via gaussian process regression.
\newblock In \emph{2020 59th IEEE Conference on Decision and Control (CDC)}, pages 860--866. IEEE, 2020.

\bibitem[Katz et~al.(2017)Katz, Barrett, Dill, Julian, and Kochenderfer]{katz2017reluplex}
Guy Katz, Clark Barrett, David~L Dill, Kyle Julian, and Mykel~J Kochenderfer.
\newblock Reluplex: An efficient {SMT} solver for verifying deep neural networks.
\newblock In \emph{International Conference on Computer Aided Verification}, pages 97--117. Springer, 2017.

\bibitem[Khan and Rue(2021)]{khan2021bayesian}
Mohammad~Emtiyaz Khan and H{\aa}vard Rue.
\newblock The bayesian learning rule.
\newblock \emph{arXiv preprint arXiv:2107.04562}, 2021.

\bibitem[K{\"o}nighofer et~al.(2020)K{\"o}nighofer, Bloem, Junges, Jansen, and Serban]{konighofer2020safe}
Bettina K{\"o}nighofer, Roderick Bloem, Sebastian Junges, Nils Jansen, and Alex Serban.
\newblock Safe reinforcement learning using probabilistic shields.
\newblock In \emph{International Conference on Concurrency Theory: 31st CONCUR 2020: Vienna, Austria (Virtual Conference)}. Schloss Dagstuhl-Leibniz-Zentrum fur Informatik GmbH, Dagstuhl Publishing, 2020.

\bibitem[Kwiatkowska et~al.(2007)Kwiatkowska, Norman, and Parker]{kwiatkowska2007stochastic}
Marta Kwiatkowska, Gethin Norman, and David Parker.
\newblock Stochastic model checking.
\newblock In \emph{International School on Formal Methods for the Design of Computer, Communication and Software Systems}, pages 220--270. Springer, 2007.

\bibitem[Lechner et~al.(2021)Lechner, {\v{Z}}ikeli{\'c}, Chatterjee, and Henzinger]{lechner2021infinite}
Mathias Lechner, {\DJ}or{\dj}e {\v{Z}}ikeli{\'c}, Krishnendu Chatterjee, and Thomas Henzinger.
\newblock Infinite time horizon safety of bayesian neural networks.
\newblock \emph{Advances in Neural Information Processing Systems}, 34, 2021.

\bibitem[Liang(2005)]{liang2005bayesian}
Faming Liang.
\newblock Bayesian neural networks for nonlinear time series forecasting.
\newblock \emph{Statistics and computing}, 15\penalty0 (1):\penalty0 13--29, 2005.

\bibitem[Liu et~al.(2019)Liu, Li, Wu, and Hsieh]{liu2018adv}
Xuanqing Liu, Yao Li, Chongruo Wu, and Cho-Jui Hsieh.
\newblock Adv-bnn: Improved adversarial defense through robust {B}ayesian neural network.
\newblock \emph{7th International Conference on Learning Representations, ICLR 2019-Conference Track Proceedings}, 2019.

\bibitem[Madry et~al.(2017)Madry, Makelov, Schmidt, Tsipras, and Vladu]{madry2017towards}
Aleksander Madry, Aleksandar Makelov, Ludwig Schmidt, Dimitris Tsipras, and Adrian Vladu.
\newblock Towards deep learning models resistant to adversarial attacks.
\newblock \emph{arXiv preprint arXiv:1706.06083}, 2017.

\bibitem[Masegosa(2020)]{masegosa2020learning}
Andres Masegosa.
\newblock Learning under model misspecification: Applications to variational and ensemble methods.
\newblock \emph{Advances in Neural Information Processing Systems}, 33:\penalty0 5479--5491, 2020.

\bibitem[McAllister and Rasmussen(2017)]{mcallister2016data}
Rowan McAllister and Carl~Edward Rasmussen.
\newblock Data-efficient reinforcement learning in continuous state-action gaussian-pomdps.
\newblock In I.~Guyon, U.~V. Luxburg, S.~Bengio, H.~Wallach, R.~Fergus, S.~Vishwanathan, and R.~Garnett, editors, \emph{Advances in Neural Information Processing Systems}, volume~30. Curran Associates, Inc., 2017.

\bibitem[Michelmore et~al.(2020)Michelmore, Wicker, Laurenti, Cardelli, Gal, and Kwiatkowska]{michelmore2020uncertainty}
Rhiannon Michelmore, Matthew Wicker, Luca Laurenti, Luca Cardelli, Yarin Gal, and Marta Kwiatkowska.
\newblock Uncertainty quantification with statistical guarantees in end-to-end autonomous driving control.
\newblock In \emph{2020 IEEE International Conference on Robotics and Automation (ICRA)}, pages 7344--7350. IEEE, 2020.

\bibitem[Mirman et~al.(2018)Mirman, Gehr, and Vechev]{mirman2018differentiable}
Matthew Mirman, Timon Gehr, and Martin Vechev.
\newblock Differentiable abstract interpretation for provably robust neural networks.
\newblock In \emph{International Conference on Machine Learning}, pages 3578--3586. PMLR, 2018.

\bibitem[Murphy(2012)]{murphy2012machine}
Kevin~P Murphy.
\newblock \emph{Machine learning: a probabilistic perspective}.
\newblock MIT press, 2012.

\bibitem[Myshkov and Julier(2016)]{myshkov2016posterior}
Pavel Myshkov and Simon Julier.
\newblock Posterior distribution analysis for bayesian inference in neural networks.
\newblock In \emph{Workshop on Bayesian Deep Learning, NIPS}, 2016.

\bibitem[Neal(2012)]{neal2012bayesian}
Radford~M Neal.
\newblock \emph{{B}ayesian learning for neural networks}, volume 118.
\newblock Springer Science \& Business Media, 2012.

\bibitem[Neal et~al.(2011)]{neal2011mcmc}
Radford~M Neal et~al.
\newblock Mcmc using hamiltonian dynamics.
\newblock \emph{Handbook of markov chain monte carlo}, 2\penalty0 (11):\penalty0 2, 2011.

\bibitem[Phan et~al.(2020)Phan, Grosu, Jansen, Paoletti, Smolka, and Stoller]{phan2020neural}
Dung~T Phan, Radu Grosu, Nils Jansen, Nicola Paoletti, Scott~A Smolka, and Scott~D Stoller.
\newblock Neural simplex architecture.
\newblock In \emph{NASA Formal Methods Symposium}, pages 97--114. Springer, 2020.

\bibitem[Polymenakos et~al.(2019)Polymenakos, Abate, and Roberts]{polymenakos2019safe}
Kyriakos Polymenakos, Alessandro Abate, and Stephen Roberts.
\newblock Safe policy search using {G}aussian process models.
\newblock In \emph{Proceedings of the 18th International Conference on Autonomous Agents and Multi Agent Systems}, pages 1565--1573. IFAAMS, 2019.

\bibitem[Polymenakos et~al.(2020)Polymenakos, Laurenti, Patane, Calliess, Cardelli, Kwiatkowska, Abate, and Roberts]{polymenakos2020safety}
Kyriakos Polymenakos, Luca Laurenti, Andrea Patane, Jan-Peter Calliess, Luca Cardelli, Marta Kwiatkowska, Alessandro Abate, and Stephen Roberts.
\newblock Safety guarantees for iterative predictions with {G}aussian processes.
\newblock In \emph{2020 59th IEEE Conference on Decision and Control (CDC)}, pages 3187--3193. IEEE, 2020.

\bibitem[Schrittwieser et~al.(2019)Schrittwieser, Antonoglou, Hubert, Simonyan, Sifre, Schmitt, Guez, Lockhart, Hassabis, Graepel, et~al.]{schrittwieser2019mastering}
Julian Schrittwieser, Ioannis Antonoglou, Thomas Hubert, Karen Simonyan, Laurent Sifre, Simon Schmitt, Arthur Guez, Edward Lockhart, Demis Hassabis, Thore Graepel, et~al.
\newblock Mastering atari, go, chess and shogi by planning with a learned model. corr abs/1911.08265 (2019).
\newblock \emph{arXiv preprint arXiv:1911.08265}, 2019.

\bibitem[Soudjani and Abate(2013)]{SA13b}
S.~Esmaeil~Zadeh Soudjani and A.~Abate.
\newblock Probabilistic reach-avoid computation for partially-degenerate stochastic processes.
\newblock \emph{IEEE Transactions on Automatic Control}, 58\penalty0 (12):\penalty0 528--534, 2013.

\bibitem[Sun et~al.(2019)Sun, Khedr, and Shoukry]{sun2019formal}
Xiaowu Sun, Haitham Khedr, and Yasser Shoukry.
\newblock Formal verification of neural network controlled autonomous systems.
\newblock In \emph{Proceedings of the 22nd ACM International Conference on Hybrid Systems: Computation and Control}, pages 147--156, 2019.

\bibitem[Sutton and Barto(1998)]{sutton}
Richard~S Sutton and Andrew~G Barto.
\newblock Reinforcement learning: An introduction, 1998.

\bibitem[Vinogradska et~al.(2016)Vinogradska, Bischoff, Nguyen-Tuong, Romer, Schmidt, and Peters]{vinogradska2016stability}
Julia Vinogradska, Bastian Bischoff, Duy Nguyen-Tuong, Anne Romer, Henner Schmidt, and Jan Peters.
\newblock Stability of controllers for gaussian process forward models.
\newblock In \emph{International Conference on Machine Learning}, pages 545--554. PMLR, 2016.

\bibitem[Wei and Liu(2021)]{wei2021safe}
Tianhao Wei and Changliu Liu.
\newblock Safe control with neural network dynamic models.
\newblock \emph{arXiv preprint arXiv:2110.01110}, 2021.

\bibitem[Wicker(2021)]{wicker2021adversarial}
Matthew Wicker.
\newblock \emph{Adversarial robustness of Bayesian neural networks}.
\newblock PhD thesis, University of Oxford, 2021.

\bibitem[Wicker et~al.(2020)Wicker, Laurenti, Patane, and Kwiatkowska]{wicker2020probabilistic}
Matthew Wicker, Luca Laurenti, Andrea Patane, and Marta Kwiatkowska.
\newblock Probabilistic safety for bayesian neural networks.
\newblock In Jonas Peters and David Sontag, editors, \emph{Proceedings of the 36th Conference on Uncertainty in Artificial Intelligence (UAI)}, volume 124 of \emph{Proceedings of Machine Learning Research}, pages 1198--1207. PMLR, 03--06 Aug 2020.

\bibitem[Wicker et~al.(2021{\natexlab{a}})Wicker, Laurenti, Patane, Chen, Zhang, and Kwiatkowska]{wicker2021bayesian}
Matthew Wicker, Luca Laurenti, Andrea Patane, Zhuotong Chen, Zheng Zhang, and Marta Kwiatkowska.
\newblock Bayesian inference with certifiable adversarial robustness.
\newblock In Arindam Banerjee and Kenji Fukumizu, editors, \emph{Proceedings of The 24th International Conference on Artificial Intelligence and Statistics}, volume 130 of \emph{Proceedings of Machine Learning Research}, pages 2431--2439. PMLR, 13--15 Apr 2021{\natexlab{a}}.

\bibitem[Wicker et~al.(2021{\natexlab{b}})Wicker, Laurenti, Patane, Paoletti, Abate, and Kwiatkowska]{wicker2021certification}
Matthew Wicker, Luca Laurenti, Andrea Patane, Nicola Paoletti, Alessandro Abate, and Marta Kwiatkowska.
\newblock Certification of iterative predictions in bayesian neural networks.
\newblock In \emph{Uncertainty in Artificial Intelligence}, pages 1713--1723. PMLR, 2021{\natexlab{b}}.

\bibitem[Wicker et~al.(2022)Wicker, Heo, Costabello, and Weller]{wicker2022robust}
Matthew~Robert Wicker, Juyeon Heo, Luca Costabello, and Adrian Weller.
\newblock Robust explanation constraints for neural networks.
\newblock In \emph{The Eleventh International Conference on Learning Representations}, 2022.

\bibitem[Wijesuriya and Abate(2019)]{bcWA19}
V.~Wijesuriya and A.~Abate.
\newblock Bayes-adaptive planning for data-efficient verification of uncertain {M}arkov decision processes.
\newblock In \emph{Proceedings of QEST, LNCS 11785}, pages 91--108, 2019.

\end{thebibliography}
\newpage
\appendix

\section{Further Experimental Details}\label{sec:appendixdetails}
%\LL{Integrate Figure in Appendix}

\iffalse
\begin{figure}[]
    \centering
    \includegraphics[width=1.0\textwidth]{Figures/Schematic.pdf}
    \caption{A diagram of the setting we consider. Data from a real cyber-physical system is recorded and a BNN model is trained on these observations. An agent acts in the environment according to a policy which is updated according to the BNN knowledge. Taking the BNN to be a faithful representation of the environment, and given a reach-avoid specification, our method verifies the policy $\pi$ against the BNN model.}
    \label{fig:diagram}
\end{figure}

We illustrate, pictorially our learning set up in Figure~\ref{fig:diagram}.
\fi

\subsection{Agent Dynamics}\label{appendix:agentdynamics}

The puck agent is derived from a classical control problem of controlling a vehicle from an initial condition to a goal state or way point \citep{Astrom08}. This scenario is more challenging than other standard benchmarks (i.e. inverted pendulum) due to both the increase state-space dimension and to the introduction of momentum which makes control more difficult. The state space of the unextended agent is a four vector containing the position in the plane as well as a vector representing the current velocity. The control signal is a two vector representing a change in the velocity (i.e. an acceleration vector). The dynamics of the puck can be given as a the following system of equations where $\eta$ determines friction, $m$ determines the mass of the puck, and $h$ determines the size of the time discretization.

$$
\dot{q} = Aq + Bc 
$$

$$
A =   \begin{bmatrix}
    1 & 0 & h & 0  \\
    0 & 1 & 0 & h  \\
    0 & 0 & 1-h\eta/m & 0  \\
    0 & 0 & 0 & 1-h\eta/m  \\
  \end{bmatrix}
$$

$$
B =   \begin{bmatrix}
    0 & 0  \\
    0 & 0  \\
    h/m & 0  \\
    0 & h/m  \\
  \end{bmatrix}
$$

The $n$ dimensional extension of the above dynamics is done by simply noting the structure of the matrices and generalizing them. In the upper-left of matrix $A$ we have the $2\times2$ identity matrix which is extended to $n \times n$. Similarly, the upper-right of $A$ is extended to $h$ times the $n \times n$ identity matrix, and the lower-right is $1-h\eta /m$ times the $n \times n$ identity matrix. For each environment and including the $n$ dimensional generalizations, time resolution, $h$, is set to 0.35, the mass of the object, $m$, is fixed to $5.0$, and the friction coefficient, $\eta$, is set to 1.0.

\subsection{Learning Parameters}

In this section we provide the hyper-parameters used for learning an initial policy and for synthesizing NN policies. In particular, we give full parameters for the environmental interaction required to learn our policies, BNN parameters to perform approximate inference, and NN parameters for neural policy synthesis.

\paragraph{Episodic Parameters}
In Table~\ref{tab:EpisodeParams} we give the parameters for our episodic learning set up. We provide the duration (number of episodes) and the amount of data collected for each episode (number of trajectories). We highlight that as this is a model-based set up, we require many fewer simulations of the system than corresponding model-free algorithms. Each environment also has an empirically tuned maximum horizon (maximum duration for each trajectory), policy size (discretization of the state-space), and obstacle aversion, $c$, as discussed in Section~\ref{Sec:Syntesis}.

\begin{table}[]\footnotesize
\centering
\begin{tabular}{lcccll}
\multicolumn{6}{c}{\textit{\textbf{Episodic Learning Parameters}}}                                                                                                                                                                          \\ \cline{2-6} 
\multicolumn{1}{l|}{}                   & \multicolumn{1}{c|}{\# Episodes} & \multicolumn{1}{c|}{\# Trajectories} & \multicolumn{1}{c|}{Max Horizon} & \multicolumn{1}{l|}{Policy Size}            & \multicolumn{1}{c|}{$c$} \\ \hline
\multicolumn{1}{|l|}{$\texttt{V1}$}     & \multicolumn{1}{c|}{15}          & \multicolumn{1}{c|}{20}              & \multicolumn{1}{c|}{25}           & \multicolumn{1}{l|}{$35 \times 35 \times 5 \times 5$} & \multicolumn{1}{l|}{0.25}   \\ \hline
\multicolumn{1}{|l|}{$\texttt{V2}$}     & \multicolumn{1}{c|}{10}          & \multicolumn{1}{c|}{15}              & \multicolumn{1}{c|}{45}           & \multicolumn{1}{l|}{$35 \times 35 \times 5 \times 5$} & \multicolumn{1}{l|}{0.25}   \\ \hline
\multicolumn{1}{|l|}{$\texttt{Zigzag}$} & \multicolumn{1}{c|}{25}          & \multicolumn{1}{c|}{15}              & \multicolumn{1}{c|}{35}           & \multicolumn{1}{l|}{$25 \times 25 \times 3 \times 3$} & \multicolumn{1}{l|}{0.125}  \\ \hline
\end{tabular}\caption{}\label{tab:EpisodeParams}
\end{table}

\paragraph{BNN Architectures}

In Table~\ref{tab:BNNParams}, we report the HMC learning parameters for our initial set up. We give details on NN architecture size, and highlight that our hidden layer uses sigmoid activation functions while the output is equipped with a linear activation function. The burn-in perior of the HMC chain is the number of samples which are automatically not included in the posterior but help initialize the chain prior to use of the Metropolis-Rosenbluth-Hastings correction step \cite{neal2012bayesian}. For each environmental set up we employ a leap-frog numerical integrator with 10 steps. The prior for all NN architectures is selected based on 2 times the variance perscribed by \cite{glorot2010understanding} which has shown to be an empirically well-performing prior in previous works \cite{wicker2020probabilistic, wicker2021bayesian}. The likelihood used to fit the BNN dynamics model  is a mean squared error ($\ell_2$) loss function.

When using variations inference, we use 1500 epochs to fit a posterior approximated by Stochastic Weight Averaging Guassian (SWAG) which does not admit a weight-space prior. We use a learning rate of 0.025 and a decay of 0.1. 

\begin{table}[]\footnotesize
\begin{tabular}{lccccccc}
\multicolumn{8}{c}{\textit{\textbf{BNN Learning Parameters}}}                                                                                                                                                                                                            \\ \cline{2-8} 
\multicolumn{1}{l|}{}                   & \multicolumn{1}{c|}{\# Layers} & \multicolumn{1}{c|}{\# Neurons} & \multicolumn{1}{c|}{Acivations} & \multicolumn{1}{c|}{Samps.} & \multicolumn{1}{l|}{Burn In} & \multicolumn{1}{c|}{LR}   & \multicolumn{1}{l|}{Decay} \\ \hline
\multicolumn{1}{|l|}{$\texttt{V1}$}     & \multicolumn{1}{c|}{1}         & \multicolumn{1}{c|}{50}         & \multicolumn{1}{c|}{sigmoid}    & \multicolumn{1}{c|}{500}           & \multicolumn{1}{c|}{25}      & \multicolumn{1}{l|}{0.05} & \multicolumn{1}{c|}{0.1}   \\ \hline
\multicolumn{1}{|l|}{$\texttt{V2}$}     & \multicolumn{1}{c|}{1}         & \multicolumn{1}{c|}{50}         & \multicolumn{1}{c|}{sigmoid}    & \multicolumn{1}{c|}{500}           & \multicolumn{1}{c|}{5}       & \multicolumn{1}{c|}{0.05} & \multicolumn{1}{c|}{0.1}   \\ \hline
\multicolumn{1}{|l|}{$\texttt{Zigzag}$} & \multicolumn{1}{c|}{1}         & \multicolumn{1}{c|}{50}         & \multicolumn{1}{c|}{sigmoid}    & \multicolumn{1}{c|}{250}           & \multicolumn{1}{c|}{15}      & \multicolumn{1}{c|}{0.1}  & \multicolumn{1}{c|}{0.1}   \\ \hline
\end{tabular}\caption{}\label{tab:BNNParams}
\end{table}

\paragraph{Neural Policy Parameters}

In our experiments, we employ an NN policy, $\pi_\theta$, comprised of a single hidden layer with 36 neurons. This is first trained to mimic actions that are sampled randomly at uniform. This training is done with SGD and is done for 15000 sampled states and actions. The NN policy is trained with 100 epochs of stochastic gradient descent with learning rate 0.00075 every time it is trained. This occurs after a BNN has been fit and used to update the actions according to loss presented in Section~\ref{Sec:Syntesis} save for no adversarial noise is taken into consideration. When we do perform synthesis, all of the same parameters are used: 15000 actions are considered and updated in parallel by SGD w.r.t. the adversarial loss defined in Section~\ref{Sec:Syntesis}.

\iffalse
\subsection{Visual Results on Learned Policies}

\begin{figure}[H]
\centering
\includegraphics[width=1.0\textwidth]{Figures/Learned_V1.pdf}
\includegraphics[width=1.0\textwidth]{Figures/Learned_V2.pdf}
\includegraphics[width=1.0\textwidth]{Figures/Learned_Zig.pdf}
\caption{\textbf{Left Column:} 200 simulated trajectories for the learned policy starting from the initial state. \textbf{Center Left Column:} A 2D visualization of the learned policy. Each arrow represents the direction of the applied force. \textbf{Center Right Column:} The epistemic uncertainty for the learned dynamics model. \textbf{Right Column:} Certified lower-bounds of probabilistic reach avoid for each abstract state according to BNN and final learned policy.}
\label{fig:LearnedSystems}
\end{figure}

In Section~\ref{sec:experiments} we provide numerical results for certification of learned policies. Here we simply visualize these results. We highlight that the uncertainty in this figure and in Figure~\ref{fig:SynthesisFigure} are identical as the BNN model is not changed, only the policy.  We find that the results on these systems are noticeably worse than those for optimal policies, but this is to be expected as worst-case guarantees are notoriously poor for naturally trained deep learning systems \cite{gowal2018effectiveness}.
\fi

\begin{figure}[H]
\centering
\includegraphics[width=1.0\textwidth]{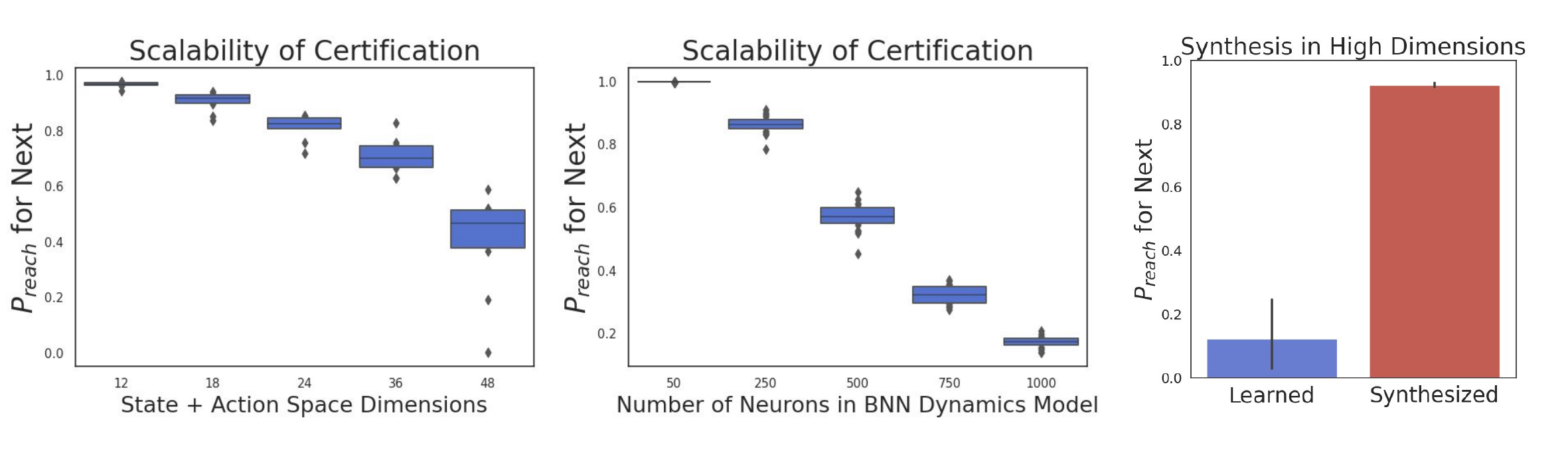}
\caption{\textbf{Left:} Certified lower bound for forward invariance with respect to an increasing number of state-space dimensions. \textbf{Center:} Certified lower bound as a function of the width of the BNN \textbf{Right:} Comparison between synthesised policy and learned policy  on the 12-dimensional puck problem.}
\label{fig:EnvScalabilityFigure}
\end{figure}

\begin{figure}[H]
\centering
\includegraphics[width=0.95\textwidth]{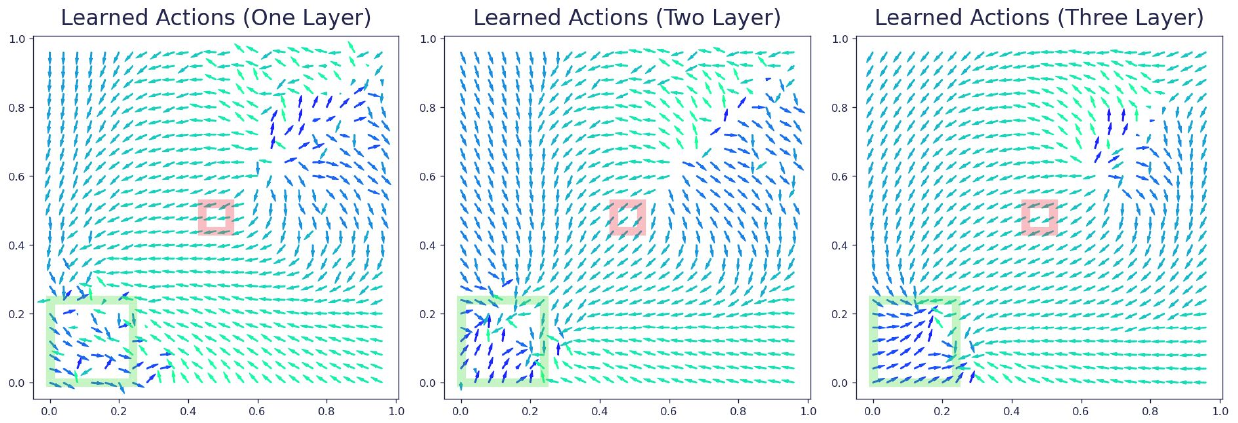}
\caption{\edit{We plot the learned policies corresponding to each of the dynamical systems whose certification is visualized in Figure~\ref{fig:DepthExperiments}. The left plot is the policy learned along with a one-layer BNN, the center plot is the policy learned along with a two-layer BNN, and the right plot is the policy learned along with a three-layer BNN.}}
\label{fig:Multilayer-Actions}
\end{figure}

\begin{figure}[H]
\centering
\includegraphics[width=0.4\textwidth]{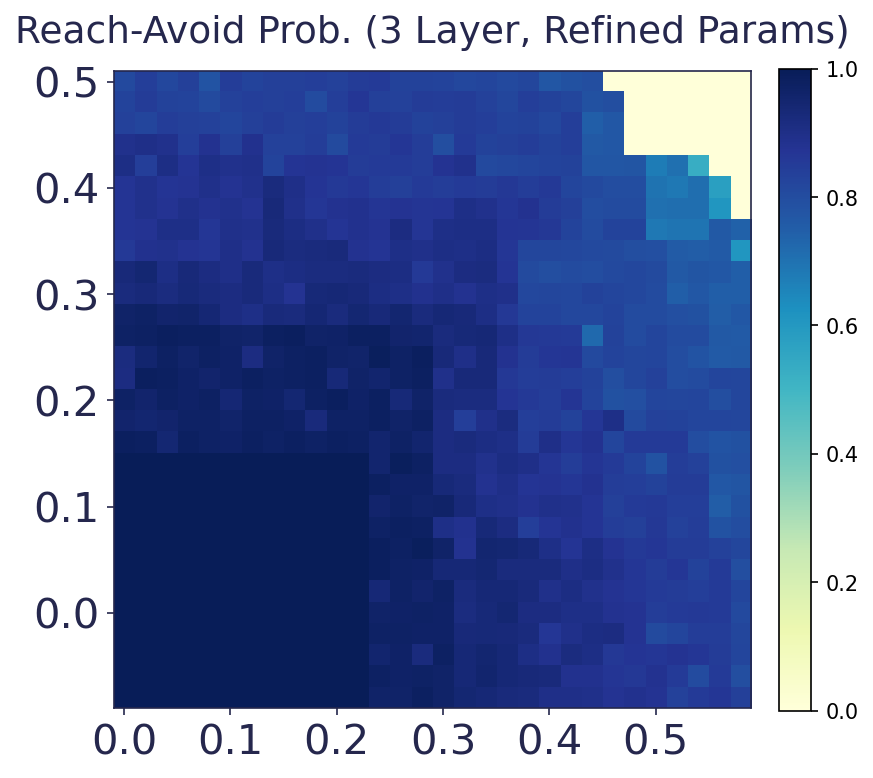}
\caption{\edit{Lower-bound reach-avoid probabilities for the three layer BNN after refining the certification parameters from the procedure in Figure~\ref{fig:DepthExperiments}. }%\ap{This discussion belong to the text. In the caption you should just say what you are plotting, i.e., certification results for the three layer BNN on the ?v1? scenario after increasing the number of BNN samples in the IBP verification.} 
}
\label{fig:3Layer-Refined}
\end{figure}

\subsection{Further Scalability Experiments}\label{app:scalability}
\edit{In this section we provide further analysis and discussion of experiments on BNN depth in our framework and provide further experiments exploring how our method scales with state-space dimensions.}

\subsubsection{Further Detail on BNN Depth Experiments}

\edit{In Figure~\ref{fig:Multilayer-Actions} we plot the policies that correspond to each of the BNNs learned for our depth experiments discussed in Section~\ref{sec:scalability} and visualized in Figure~\ref{fig:DepthExperiments}. We highlight that each of the policies are qualitatively very similar, though they may have slight quantitative differences. In Figure~\ref{fig:3Layer-Refined} we plot the result of the more computationally expensive certification on the three-layer BNN. Though our method struggles to get strong certification for the system with a three-layer BNN in Figure~\ref{fig:DepthExperiments}, by tuning the certification parameters we are able to get a much tighter lower-bound (average lower-bound safety probability $0.621 \rightarrow 0.867$). We notice that many of the previously uncertified (i.e., lower-bound probability 0.0) states have lower-bounds above 0.6 when more BNN samples are used in the certification procedure.} 

\subsubsection{Scaling with State-Space Dimensionality}
In this section, we evaluate how our method performs while varying the dimensionality of the environment and the size of the BNN architecture.
In particular we perform the analyses using an $n$-dimensional generalizations of the \texttt{v1} layout - described by $3n$ continuous values, $n$ dimensions for position, velocity, and action spaces, respectively.
For such high-dimensional state space, full discretisation of the state space becomes infeasible.
In order to overcome this, we consider a forward-invariance variant of the reach-avoid property from our previous experiments, where the agent goal is iteratively moved at each step in order to guide it to the global goal. \edit{In other words, here we consider one-step reachability ($N=1$), which allows us to significantly reduce the set of discretised states to consider (by restricting to neighbour states that can be reached in one step only).} %\np{next sentence is redundant. can we replace it with ``In other words, here we consider one-step reachability ($N=1$), which allows us to significantly reduce the set of discretised states to consider (by restricting to neighbour states that can be reached in one step only).''?} So that the forward invariance property  requires that the agent move towards the goal in the next time step while avoiding the obstacles, at each time step.

The results for these analyses are given in Figure~\ref{fig:EnvScalabilityFigure}.
The left plot of the figure, shows how even for 48 dimensions we are still able to obtain non-vacous bounds at $0.5$, but as expected, the quality of the bound decreases quickly with the size of the state space and actions.
The centre plot depicts the certified bounds obtained for an increasing number of BNN hidden units, up until $1000$ for the 12-dimensional \texttt{v1} scenario. 
Finally, the right plot of Figure~\ref{fig:EnvScalabilityFigure} shows that  our synthesis algorithm strongly improves on the initially learned policy, even in higher-dimensional settings. We accomplish this improvement by following the neural policy synthesis method presented in Section~\ref{Sec:Syntesis} where the worst-case $\epsilon$ for tuning our actions is set to 0.025. 

\edit{This analysis of the forward invariance property allows us to understand how our algorithm, particularly the evaluation of the $R$ function, scales to larger NNs and state spaces. However, for state-space dimensions that are greater than the ones analyzed in the prior section of this paper, certification of the entire state-space is computationally infeasible due to the exponential nature of the discretization involved.}

\section{Proofs}

\textbf{Proof of Proposition \ref{Prob:verification}}
In what follows, we omit $\pi$ (which is given and held constant) from the probabilities for a more compact notation. 
The proof is by induction. The base case is $k=N$, for which we have
$$V_N^{\pi}(x) =\mathbf{1}_\mathrm{G}(x)= P_{reach}(\mathrm{G},\mathrm{S},x,[N,N]),  $$
which holds trivially. Under the assumption that, for any given $k\in [0,N-1]$, it holds that 
\begin{align}
    \label{Eqn:IndBaseCase}
  V_{k+1}^{\pi}(x) = P_{reach}(\mathrm{G},\mathrm{S},x,[k+1,N]),
\end{align}
we show the induction step for time step $k$. 
In particular, 
\begin{align*}
     &P_{reach}(\mathrm{G},\mathrm{S},x,[k,N]|\pi)=\\
     &Pr( \mathbf{x}_k \in \mathrm{G}  |\mathbf{x}_k = x) +\sum_{j=k+1}^{N}Pr( \mathbf{x}_j \in \mathrm{G} \wedge \forall j' \in [k,j) , \mathbf{x}_{j'} \in \mathrm{S} |\mathbf{x}_k= x)= \\
     &\mathbf{1}_{\mathrm{G}}(x) + \mathbf{1}_{\mathrm{S}}(x)\sum_{j=k+1}^{N}Pr( \mathbf{x}_j \in \mathrm{G} \wedge \forall j' \in [k,j), \mathbf{x}_{j'} \in \mathrm{S} |\mathbf{x}_k= x)
\end{align*}     
Now in order to conclude the proof we want to show that
\begin{align*}
    \sum_{j=k+1}^{N}Pr( \mathbf{x}_j \in \mathrm{G} \wedge \forall j' \in [k,j)+1, \mathbf{x}_{j'} \in \mathrm{S} |&\mathbf{x}_k= x) =\\
    & \int V_{k+1}^{\pi}(\bar{x})p(\bar{x} \mid (x,\pi_k(x)),\mathcal{D} ) d\bar{x}.
\end{align*}  
This can be done as follow
\begin{align*}
   & \sum_{j=k+1}^{N}Pr( \mathbf{x}_j \in \mathrm{G} \wedge \forall j' \in [k+1,j), \mathbf{x}_{j'} \in \mathrm{S} |\mathbf{x}_k= x)=\\
 & Pr( \mathbf{x}_{k+1} \in \mathrm{G} |\mathbf{x}_k=x)+ \\
 &\quad \sum_{j=k+2}^{N}Pr( \mathbf{x}_j \in \mathrm{G} \wedge \forall j' \in [k+1,j), \mathbf{x}_{j'} \in \mathrm{S} |  \mathbf{x}_k= x)=\\
  %& \int_{\mathrm{G}} p(\bar{x} \mid (x,\pi_k(x)),\mathcal{D} )d\bar{x} + \\
% &\quad \sum_{j=k+2}^{N}Pr( \mathbf{x}_j \in \mathrm{G} \wedge \forall j' \in [k+1,j), \mathbf{x}_{j'} \in \mathrm{S} |\rho(k+1)\in \mathrm{S},  \mathbf{x}_k= x)Pr(\rho(k+1)\in \mathrm{S} | \mathbf{x}_k= x)=\\
   & \int_{\mathrm{G}} p(\bar{x} \mid (x,\pi_k(x)),\mathcal{D} )d\bar{x} +\\
   &\quad \sum_{j=k+2}^{N}\int_{\mathrm{S}}Pr( \mathbf{x}_j \in \mathrm{G} \wedge \forall j' \in [k+2,j), \mathbf{x}_{j'} \in \mathrm{S} \wedge \mathbf{x}_{k+1}=\bar{x} |  \mathbf{x}_k= x) d\bar{x}=\\
  & \int_{\mathrm{G}} p(\bar{x} \mid (x,\pi_k(x)),\mathcal{D} )d\bar{x} +\\
  & \quad \sum_{j=k+2}^{N}\int_{\mathrm{S}} Pr( \mathbf{x}_j \in \mathrm{G} \wedge \forall j' \in [k+2,j), \mathbf{x}_{j'} \in \mathrm{S}   | \mathbf{x}_{k+1}=\bar{x} )p(\bar{x} \mid (x,\pi_k(x)),\mathcal{D} )d{\bar{x}}=\\
   & \int \big( \mathbf{1}_{\mathrm{G}}(\bar{x}) +\\
   &\quad \mathbf{1}_{\mathrm{S}}(\bar{x}) \sum_{j=k+2}^{N} Pr( \mathbf{x}_j \in \mathrm{G} \wedge \forall j' \in [k+2,j), \mathbf{x}_{j'} \in \mathrm{S}   | \mathbf{x}_{k+1}=\bar{x} )\big)p(\bar{x} \mid (x,\pi_k(x)),\mathcal{D} )d{\bar{x}}=\\
    & \int V_{k+1}^{\pi}(\bar{x})p(\bar{x} \mid (x,\pi_k(x)),\mathcal{D} )d{\bar{x}}\\
\end{align*}
where the third step holds by application of Bayes rule over multiple events.

\textbf{Proof of Theorem \ref{th:VerificationLoerBound}}
The proof is by induction. The base case is $k=N$, for which we have
$$\inf_{x\in q}V_N^{\pi}(x)=\inf_{x\in q}\mathbf{1}_{\mathrm{G}}(x)=\mathbf{1}_{\mathrm{G}}(q)=K_N^{\pi}(q). $$
Next, under the assumption that for any $k\in \{0,N-1 \}$ it holds that 
\begin{align*}
    %\label{Eqn:IndBaseCase}
 \inf_{x\in q} V_{k+1}^{\pi}(x) \geq K_{k+1}^{\pi}(q),
\end{align*}
we can work on the induction step: 
in order to derive it, it is enough to show that for any $\epsilon >0$
\begin{align*} 
&\int  V_{k+1}^{\pi}(\bar{x})p(\bar{x} \mid (x,\pi_k(x)),\mathcal{D} )d\bar{x} \geq \\ &\quad F([-\epsilon,\epsilon]|\sigma^2)^n \sum_{i=1}^{n_p}\int_{H^{q,\pi}_{k,i}}  v_{i-1}  p_{\mathbf{w}}(w|\mathcal{D})  dw,
\end{align*}
where $F([-\epsilon,\epsilon]|\sigma^2)=\text{erf}(\frac{\epsilon}{\sqrt{2 \sigma^2}})$ is the cumulative function distribution for a normal random variable with zero mean and variance $\sigma^2$ being within $[-\epsilon,\epsilon].$
This can be argued by rewriting the first term in parameter space (recall that the stochastic kernel $T$ is induced by $p_{\mathbf{w}}(w|\mathcal{D})$) and providing a lower bound, as follows: 
\begin{align*}
    & \int  V_{k+1}^{\pi}(\bar{x})p(\bar{x} \mid (x,\pi_k(x)),\mathcal{D} )d\bar{x}= \\ 
    & \text{(By definition of predictive distribution)}  \\
    & \int \big( \int  V_{k+1}^{\pi}(\bar{x}) p(\bar{x}|(x,u),w) d\bar{x} \big) p_{\mathbf{w}}(w|\mathcal{D})  dw \geq \\ 
     & \text{(By $V_{k+1}^k$ being non negative everywhere and by the Gaussian likelihood)}  \\
     & \int \big( \int_{f^{w}(x,\pi(x,k))+\epsilon}^{f^{w}(x,\pi(x,k)-\epsilon}  V_{k+1}^{\pi}(\bar{x})\mathcal{N}(\bar{x}|f^{w}(x,\pi(x,k)),\sigma^2\cdot I) d\bar{x} \big) p_{\mathbf{w}}(w|\mathcal{D})  dw\geq \\ 
       & \text{(By standard inequalities of integrals)}  \\
  & \int  \inf_{\bar{\gamma}\in [-\epsilon,\epsilon]}  V_{k+1}^{\pi}(f^{w}(x,\pi(x,k)+\bar{\gamma}) \big( \int_{[-\epsilon,\epsilon]^n} \mathcal{N}(\gamma|0,\sigma^2) d\gamma \big)^n p_{\mathbf{w}}(w|\mathcal{D})  dw\geq \\  
       & \text{(By the assumptions that for $i\neq j$ $H^{q,\pi}_{k,i}$ and $H^{q,\pi}_{k,j}$ are non-overlapping)}  \\
    &\big( \int_{[-\epsilon,\epsilon]} \mathcal{N}(\gamma|0,\sigma^2) d\gamma \big)^n \sum_{i=1}^{n_p}\int_{H^{q,\pi,\epsilon}_{k,i}}  \inf_{\bar{\gamma}\in [-\epsilon,\epsilon]}  V_{k+1}^{\pi}(f^{w}(x,\pi(x,k)+\bar{\gamma})   p_{\mathbf{w}}(w|\mathcal{D})  dw,\\
       & \text{(By the fact that $v_i\leq \inf_{x\in q} V_{k+1}^{\pi}(f^{w}(x,\pi(x,k)+\bar{\gamma})$ )}  \\
    &\big( \int_{[-\epsilon,\epsilon]} \mathcal{N}(\gamma|0,\sigma^2) d\gamma \big)^n \sum_{i=1}^{n_p}v_i \int_{H^{q,\pi,\epsilon}_{k,i}}   p_{\mathbf{w}}(w|\mathcal{D})  dw,
\end{align*}
     where the last step concludes the proof  because, by the induction hypothesis, we know that for $q' \subseteq \mathbb{R}^n$
     $$
\inf_{\bar{x}\in q'}     V_{k+1}^{\pi}(\bar{x})\geq K_{k+1}^\pi(q')
     $$ 
     and by the construction of sets $H^{q,\pi}_{k,i}$ for each of its weights  $K_{k+1}^\pi(f^w(\bar{x},\pi(x,k))$ is lower bounded by $v_{i-1}$.

\end{document}